\pdfoutput=1

\documentclass[letterpaper, 10 pt, conference]{ieeeconf}  
\IEEEoverridecommandlockouts                              

\usepackage{theorem}
\usepackage{enumerate}

\usepackage{tikz}
\usetikzlibrary{fit,calc}
\usepackage{relsize}
\newcommand{\tikzmark}[1]{\tikz[overlay,remember picture,baseline] \node [anchor=base] (#1) {};}

\usepackage[colorlinks]{hyperref}
\hypersetup{
   bookmarksnumbered,
   pdfstartview={FitH},
   citecolor={black},
   linkcolor={black},
   urlcolor={black},
   pdfpagemode={UseOutlines}
}

\usepackage[colorlinks]{hyperref}

\hypersetup{
   bookmarksnumbered,
   pdfstartview={FitH},
   citecolor={black},
   linkcolor={black},
   urlcolor={black},
   pdfpagemode={UseOutlines}
}

\usepackage{dsfont}
\usepackage{placeins}
\usepackage{color}
\usepackage{mathtools}
\usepackage{url}
\usepackage{exscale}
\usepackage{subfigure}
\usepackage[font=footnotesize]{caption}
\usepackage{array}
\usepackage{epsfig}
\usepackage{xspace}
\usepackage{bm}
\usepackage{amsmath}

\usepackage{amssymb}
\usepackage{stmaryrd}
\usepackage{comment}
\usepackage{fancybox}
\usepackage{multirow}
\usepackage[vlined, linesnumbered, boxruled]{algorithm2e}
\usepackage{relsize}
\usepackage{rotating}

\usepackage{cite}
\usepackage{multicol}

\usepackage{etoolbox}

\newtoggle{todo}

\definecolor{gray}{RGB}{190,190,190}
\definecolor{darkgreen}{rgb}{0,0.5,0} 
\definecolor{fullred}{rgb}{0.85,.0,.1} 
\definecolor{brown}{rgb}{0.65,0.16,0.16}

\newcommand{\AlgRRT}{\ensuremath{{\mathrm{RRT}}}}

\newcommand{\AlgRRG}{\ensuremath{{\mathrm{RRG}}}}

\newcommand{\AlgRRTstar}{\ensuremath{{\mathrm{RRT}^*}}}
\newcommand{\AlgRRTsharp}{\ensuremath{{\mathrm{RRT}^{\tiny \#}}}}
\newcommand{\AlgClosedLoopRRTsharp}{\ensuremath{{\mathrm{CL\textbf{-}RRT}^{\tiny \#}}}}
\newcommand{\AlgClosedLoopRRT}{\ensuremath{{\mathrm{CL\textbf{-}RRT}}}}

\newcommand{\reals}{\mathbb{R}}  
\newcommand{\naturals}{\mathbb{N}} 	

\newcommand{\graph}[1]{\mathcal{#1}}		

\newcommand{\True}{{\tt True}}
\newcommand{\False}{{\tt False}}

\newcommand{\outgoingEdgeList}{\mathtt{outgoing}}			
\newcommand{\tailRefNode}{\mathtt{tail}}			
\newcommand{\headRefNode}{\mathtt{head}}			
\newcommand{\referenceTrajectoryValue}{r}			

\newcommand{\vCtocBarValue}{\mathtt{\bar{g}}}	
\newcommand{\vCtocValue}{\mathtt{g}}	
\newcommand{\vCtocStarValue}{\mathtt{g^{*}}}
\newcommand{\vHValue}{\mathtt{h}}			

\newcommand{\PrcSample}{\mathtt{Sample}} 
\newcommand{\PrcExtend}{\mathtt{Extend}} 
\newcommand{\PrcNearest}{\mathtt{Nearest}} 
\newcommand{\PrcNear}{\mathtt{Near}}      
\newcommand{\PrcSteer}{\mathtt{Steer}}   
\newcommand{\PrcObstacleFree}{\mathtt{ObstacleFree}} 
\newcommand{\PrcReplan}{\mathtt{Replan}} 

\newcommand{\PrcSuccessor}{\mathtt{succ}} 
\newcommand{\PrcPredecessor}{\mathtt{pred}} 

\newcommand{\hideMaterial}[1]{}
\newcommand{\hideoldproofofnegresult}[1]{ }

\newcommand{\vYRand}{y_{\mathrm{rand}}}
\newcommand{\vYFrom}{y_{\mathrm{from}}}
\newcommand{\vYTo}{y_{\mathrm{to}}}
\newcommand{\vYPrime}{y^{\prime}}
\newcommand{\vClTX}{\mathcal{T}_{x}}
\newcommand{\vY}{y}
\newcommand{\vCVY}{V_{y}}
\newcommand{\vCEY}{E_{y}}

\newcommand{\vClQ}{\mathcal{Q}}

\newcommand{\vClS}{\mathcal{S}}
\newcommand{\vSigma}{\sigma}
\newcommand{\vYInit}{y_{\mathrm{init}}}
\newcommand{\vXInit}{x_{\mathrm{init}}}

\newcommand{\vClQGoal}{\mathcal{Q}_{\mathrm{goal}}}

\newcommand{\vClY}{\mathcal{Y}}

\newcommand{\vCY}{Y}
\newcommand{\vCYFree}{Y_\mathrm{free}}
\newcommand{\vCYGoal}{Y_{\mathrm{goal}}}

\newcommand{\vClGY}{\mathcal{G}_{y}}
\newcommand{\vClGSigma}{\mathcal{G}_{\sigma}}

\newcommand{\vParentTrajectory}{\mathtt{p}_{\sigma}}
\newcommand{\vParentReference}{\mathtt{p}_{y}}

\newcommand{\vVY}{v_{y}}
\newcommand{\vVSigma}{v_{\sigma}}

\newcommand{\vEY}{e_{y}}
\newcommand{\vESigma}{e_{\sigma}}

\newcommand{\vTail}{\mathtt{tail}}
\newcommand{\vHead}{\mathtt{head}}
\newcommand{\vOutgoing}{\mathtt{outgoing}}

\newcommand{\PrcOutputMap}{\mathtt{OutputMap}}
\newcommand{\PrcInitialize}{\mathtt{Initialize}}
\newcommand{\PrcConstructSolution}{\mathtt{ConstrSolution}}

\newcommand{\PrcTopKey}{\mathtt{top\_key}}

\newcommand{\PrcPop}{\mathtt{pop}}
\newcommand{\PrcOutgoing}{\mathtt{outgoing}}
\newcommand{\PrcIncoming}{\mathtt{incoming}}

\newcommand{\PrcUpdate}{\mathtt{update}}
\newcommand{\PrcInsert}{\mathtt{insert}}
\newcommand{\PrcRemove}{\mathtt{remove}}

\newcommand{\vClX}{\mathcal{X}}
\newcommand{\vClU}{\mathcal{U}}

\newcommand{\vCX}{X}
\newcommand{\vCU}{U}
\newcommand{\vCXFree}{X_{\mathrm{free}}}
\newcommand{\vCXObs}{X_{\mathrm{obs}}}
\newcommand{\vCXGoal}{X_{\mathrm{goal}}}

\SetKwFunction{fRRTsharp}{\textbf{CL-RRT$_{}^{\tiny \#}$}}

\renewcommand{\PrcSample}{\mathtt{Sample}}
\SetKwFunction{fSample}{$\PrcSample$}

\SetKwFunction{fExtend}{Extend}
\SetKwFunction{fReplan}{Replan}
\SetKwFunction{fParent}{parent}
\SetKwFunction{fFront}{front}
\SetKwFunction{fBack}{back}
\SetKwFunction{fFind}{find}
\SetKwFunction{fConstructSolution}{ConstrSolution}

\newcommand{\PrcOutputNode}{\mathtt{OutNode}}
\newcommand{\PrcOutputEdge}{\mathtt{OutEdge}}
\newcommand{\PrcTrajectoryNode}{\mathtt{TrajNode}}
\newcommand{\PrcTrajectoryEdge}{\mathtt{TrajEdge}}

\SetKwFunction{fInitStateNode}{StateNode}
\SetKwFunction{fInitStateEdge}{StateEdge}
\SetKwFunction{fInitOutputNode}{OutNode}
\SetKwFunction{fInitOutputEdge}{OutEdge}
\SetKwFunction{fInitTrajectoryNode}{TrajNode}
\SetKwFunction{fInitTrajectoryEdge}{TrajEdge}

\SetKwFunction{fInitialize}{Initialize}
\SetKwFunction{fOutputMap}{OutputMap}
\SetKwFunction{fComputeHeuristic}{ComputeHeuristic}
\SetKwFunction{fNearest}{Nearest}
\SetKwFunction{fSteer}{Steer}
\SetKwFunction{fObstacleFree}{ObstacleFree}
\SetKwFunction{fKey}{Key}
\SetKwFunction{fUpdateQueue}{UpdateQueue}
\SetKwFunction{fUpdateGoal}{UpdateGoal}
\SetKwFunction{fNear}{Near}
\SetKwFunction{fLine}{Line}
\SetKwFunction{fBegin}{begin}
\SetKwFunction{fEnd}{end}
\SetKwFunction{fFront}{front}
\SetKwFunction{fBack}{back}
\SetKwFunction{fTail}{tail}
\SetKwFunction{fHead}{head}
\SetKwFunction{vSucc}{succ}  
\SetKwFunction{fG}{g}
\SetKwFunction{fCostValue}{Cost}
\SetKwFunction{fLMC}{lmc}
\SetKwFunction{fLMCPrime}{lmc$^{\prime}$}
\SetKwFunction{fParentTrajectory}{$\sigma_{\mathrm{parent}}$}

\SetKwFunction{fIncludeVertex}{IncludeVertex}
\SetKwFunction{fPropagate}{Propagate}
\SetKwFunction{fValue}{value}
\SetKwFunction{fComputeHeuristic}{ComputeHeuristic}	
\SetKwFunction{fTop}{top}
\SetKwFunction{fTopKey}{top\_key}
\SetKwFunction{fPop}{pop}
\SetKwFunction{fOutgoing}{outgoing} 
\SetKwFunction{fSuccessor}{succ}
\SetKwFunction{fPredecessor}{succ}
\SetKwFunction{fUpdate}{update}
\SetKwFunction{fInsert}{insert}   
\SetKwFunction{fRemove}{remove}

	\newcommand{\vZ}{z}
	\newcommand{\vCZ}{Z}
	\newcommand{\vVZ}{v_{z}}
	\newcommand{\vCVZ}{V_{z}}
	\newcommand{\vV}{v}
	\newcommand{\vCVNear}{V_{\mathrm{near}}}
	\newcommand{\vVYNearest}{v_{y,\mathrm{nearest}}}
	\newcommand{\vVYNew}{v_{y,\mathrm{new}}}

    \newcommand{\vSigmaX}{\sigma_{x}}	 
	\newcommand{\vSigmaY}{\sigma_{y}}

    \newcommand{\vVSigmaSucc}{v_{\sigma,\mathrm{succ}}}
    \newcommand{\vVSigmaPred}{v_{\sigma,\mathrm{pred}}}
    
    \newcommand{\vVSigmaNew}{v_{\sigma,\mathrm{new}}}
    \newcommand{\vRNew}{r_{\mathrm{new}}}
	\newcommand{\vYNew}{y_{\mathrm{new}}} 
	\newcommand{\vCEYPred}{E_{y,\mathrm{pred}}}
    \newcommand{\vCEYSucc}{E_{y,\mathrm{succ}}}
    \newcommand{\vVYPred}{v_{y,\mathrm{pred}}}
    \newcommand{\vVYSucc}{v_{y,\mathrm{succ}}}
    \newcommand{\vXPred}{x_\mathrm{pred}} 
    \newcommand{\vR}{r} 
    \newcommand{\vX}{x} 

\title{\LARGE \bf
Sampling-based Algorithms for Optimal Motion Planning \\Using Closed-loop Prediction
}

\author{Oktay Arslan$^{1}$\thanks{$^{1}$Oktay Arslan is a Robotics, PhD Candidate with the D. Guggenheim School
of Aerospace Engineering and the Institute for Robotics and Intelligent Machines at the Georgia Institute of Technology, Atlanta, GA 30332, USA, Email:{\rm\small oktay@gatech.edu}. He performed this research while at Mitsubishi Electric Research Laboratories, Cambridge,
MA 02139, USA.} 
\and 
Karl Berntorp$^{2}$\thanks{$^{2}$Karl Berntorp is with Mitsubishi Electric Research Laboratories, Cambridge,
MA 02139, USA, Email:{\rm\small karl.o.berntorp@ieee.org}.}
\and
Panagiotis Tsiotras$^{3}$\thanks{$^{3}$Panagiotis Tsiotras is with the faculty of D. Guggenheim School of Aerospace Engineering and the Institute for Robotics and Intelligent Machines at the Georgia Institute of Technology, Atlanta, GA 30332-0150, USA, Email: {tsiotras@gatech.edu}.}
}

\begin{document}

\maketitle
\thispagestyle{empty}
\pagestyle{empty}

\begin{abstract}

Motion planning under differential constraints, \textit{kinodynamic motion planning}, is one of the canonical problems in robotics. Currently, state-of-the-art methods evolve around kinodynamic variants of popular sampling-based algorithms, such as Rapidly-exploring Random Trees (RRTs). However, there are still challenges remaining, for example, how to include complex dynamics while guaranteeing optimality. If the open-loop dynamics are unstable, exploration by random sampling in control space becomes inefficient. 
We describe a new sampling-based algorithm, called $\AlgClosedLoopRRTsharp$, which leverages ideas from the $\AlgRRTsharp$ algorithm and a variant of the $\AlgRRT$ algorithm that generates trajectories using closed-loop prediction. The idea of planning with closed-loop prediction allows us to handle complex unstable dynamics and avoids the need to find computationally hard steering procedures. The search technique presented in the \AlgRRTsharp{} algorithm allows us to improve the solution quality by searching over alternative reference trajectories. Numerical simulations using a nonholonomic system demonstrate the benefits of the proposed approach.

\end{abstract}

\section{Introduction} \label{section:introduction}

Motion planning is ubiquitous in many applications where different levels of autonomy is desired.  Loosely speaking, given a system that is subject to a set of differential constraints, an initial state, a final state, a set of obstacles, and a goal region, the \textit{motion-planning problem} is to find a control input that drives the system from its initial state to the goal region. This problem is computationally hard to solve \cite{reif1979complexity}.

One approach to solve the motion-planning problems is to divide the problem into two subproblems: path planning and path tracking. The main drawback of this approach is lack of dynamic feasibility guarantees. Still, it has been successfully applied to robotic applications in which the underlying system has redundant control authority (e.g., robotic manipulators). 
 Another  class of algorithms is randomized planners, which solve  the motion-planning problem in a single step. Notably, the kinodynamic version of  Rapidly-Exploring Random Tree (\AlgRRT{})  incrementally grows a tree of trajectories in the state space by sampling  control inputs and simulating the motion of the system with these random control inputs over a time horizon~\cite{lavalle2006planning,lavalle2001randomizedkin}. Hence, the trajectories that are generated by  \AlgRRT{}  are dynamically feasible by construction. Recently,  \AlgRRT{}  and its variants were successfully applied to  robotic systems~\cite{kuffner2002dynamically,leonard2008perception} and different classes of stochastic problems~\cite{arslan2014information}. Unlike standard \AlgRRT{}, these variants were usually implemented  to compute a solution quickly and improve it in the remaining time until the execution of the motion plan. However, \AlgRRT{}  computes suboptimal solutions~\cite{KaramanFra2011}.

One drawback with kinodynamic \AlgRRT{} is that exploration via random selection of control inputs is  inefficient when the dynamics are complex and/or unstable. To remedy this, \cite{kuwata2008motion} proposed  \AlgClosedLoopRRT{}, which uses closed-loop prediction for trajectory generation. Instead of sampling in the control space, the proposed approach grows a tree in the reference space. Each path of the tree represents a reference trajectory that acts as an input to the closed-loop system. The desired behaviors of the system are prescribed as specifications for a  controller that is used to track a given reference trajectory. Each edge of the tree is associated with a segment of a reference trajectory and a state trajectory of the system, computed by closed-loop prediction. 

Several papers address the   suboptimality of \AlgRRT{}. In~\cite{KaramanFra2011}, an algorithm with asymptotic optimality guarantee, \AlgRRTstar{}, was developed.   \AlgRRTstar{}  has been extended to solve motion planning problems under differential constraints~\cite{karaman2011anytime,karaman2010optimal}. The proposed algorithms are asymptotically optimal when a steering procedure that satisfies certain conditions is provided. However, developing efficient steering procedures that solve point-to-point motion planning, essentially  a two-point boundary value problem, is generally hard \cite{vinter2010optimal}.

Here, we propose a new asymptotically optimal motion-planning algorithm,  \AlgClosedLoopRRTsharp{}, by leveraging ideas from the \AlgClosedLoopRRT{}~\cite{kuwata2008motion} and the \AlgRRTsharp{} algorithms~\cite{arslan2013useofrelaxation,arslan2015dynamic,arslan2015dp}. To handle differential constraints, instead of sampling  in the control space, our approach samples in the output space and incrementally grows a graph whose edges correspond to segments of reference trajectories. The algorithm also keeps another graph to store state trajectories of the closed-loop system when it is inputed with a certain path in the graph of reference trajectories. Hence, we avoid the need for complicated steering procedures and the resulting trajectory satisfies the differential constraints by construction. To improve the solution quality,  \AlgClosedLoopRRTsharp{}  searches among alternative paths of the graph of reference trajectories. The proposed algorithm checks different reference trajectories and simulates the system  forward in time, as needed. Finally, the algorithm provides the segments of reference trajectories that yield the lowest-cost state trajectory of the closed-loop system.

\section{Problem Formulation}\label{section:problem_formulation}


Let $\vCX \subseteq \reals^{n}$, $\vCY \subseteq \reals^{p}$ and $\vCU \subseteq \reals^{m}$ be compact sets. We assume that the system dynamics can be described by a nonlinear differential equation of the form
\begin{align}
\dot{x}(t) =& f(x(t),u(t)), \quad	x(0) = x_{0}, \nonumber\\
y(t) =& h(x(t),u(t)),
\end{align}
where the system state $x(t) \in \vCX$, the system output $y(t) \in \vCY$, the control $u(t) \in \vCU$, for all $t, x_{0} \in \vCX$, and $f$ and $h$ are smooth (continuously differentiable) functions describing the time evolution of the system dynamics. Let $\vClX$ denote the set of all essentially bounded measurable functions mapped from $[0,T]$ to $\vCX$ for any $T \in \reals_{>0}$ and define $\vClY$ and $\vClU$ similarly. The functions in $\vClX$, $\vClY$, and $\vClU$ are called \textit{state trajectories}, \textit{output trajectories}, and \textit{controls}, respectively. 

Let $\vCXObs$ and $\vCXGoal$, called the \textit{obstacle space} and the \textit{goal region}, be open subsets of $\vCX$. Let $\vCXFree$, also called the \textit{free space}, denote the set defined as $\vCX \setminus \vCXObs$.

The smooth function $h$ describes the output $y$ that we wish to control. Loosely speaking, we are particularly interested in the class of control problems in which we wish to track a time-varying reference trajectory $r(t)$.  called the \textit{trajectory-generation} problem. We assume that given a desired output value $y^{\prime} \in \vCY$, and a current output value $y \in \vCY$ of the system, the control law $\phi: (y^{\prime},y) \mapsto u \in \vCU$ computes a control input such that the closed-loop simulation of the system yields a good tracking performance as time evolves.

\subsection{Problem Statement}


Given the state space $\vCX$, obstacle region $\vCXObs$, goal region $\vCXGoal$, and smooth functions $f$ and $h$ that describe the system dynamics, find a reference trajectory $r \in \vClY$ with domain $[0,T]$ for some $T \in \reals_{>0}$ such that the corresponding unique state trajectory $x \in \vClX$, output trajectory $y \in \vClY$, and control $u \in \vClU$ that are computed by closed-loop simulation,
\begin{itemize}
\item obeys the differential constraints, 
\begin{align*}
\dot{x}(t) &= f(x(t),u(t)) \quad	x(0) = x_{0},\\
 y(t) &= h(x(t),u(t)) \text{~for all~} t \in [0,T],
\end{align*}

\item avoids the obstacles, i.e., $x(t) \in \vCXFree$ for all $t \in [0,T]$,
\item reaches the goal region, i.e., $x(T) \in \vCXGoal$,
\item and minimizes ${}J(x,u,r) = \int_{0}^{T} g(x(t),u(t), r(t))\, \mathrm{d}t$
\end{itemize}

\subsection{Primitive Procedures}

Following are the definitions of the primitive procedures used by the \AlgClosedLoopRRTsharp{} algorithm (for details, see~\cite{KaramanFra2011}).

\textit{Sampling:} $\PrcSample : \omega \mapsto  \left\{ \PrcSample_{i}(\omega) \right\}_{i \in \naturals_{0}} \subset \vCYFree$ returns independent and identically distributed (i.i.d.) samples $\PrcSample_{i}, \, i \in \naturals_{0	}$ from $\vCYFree$. 

\textit{Nearest Neighbor:} Given a graph $\vClGY = (\vCVY, \vCEY)$, where $\vCVY \in \vCY$, a point $\vY \in \vCY$, the function $\PrcNearest: (\vClGY, \vY) \mapsto \vVY \in \vCVY$ returns the node in $\vCVY$ that is ``closest'' to $\vY$ in terms of a given distance function. We use the  Euclidean distance. 

\textit{Near Neighbors:} Given a graph $\vClGY = (\vCVY, \vCEY)$, where $\vCVY \in \vCY$, a point $\vY \in \vCY$, and a positive real number $d \in \reals_{>0}$, the function $\PrcNearest: (\vClGY, \vY, d) \mapsto \vVY \in \vCVY^{\prime} \subset \vCVY$ returns the nodes in $\vCVY$ that are contained in a ball of radius $d$ centered at $\vY$. 

\textit{Steering:} Given two points $\vYFrom, \vYTo \in \vCY$, the function $\PrcSteer: (\vYFrom, \vYTo) \mapsto \vYPrime$ returns a point $\vYPrime \in \vCY$ such that $\vYPrime$ is ``closer'' to $\vYTo$ than $\vYFrom$ is. In this work, the point $\vYPrime$ returned by the function $\PrcSteer$ will be such that $\vYPrime$ minimizes $\|\vYPrime - \vYTo\|$ while at the same time maintaining $\|\vYPrime - \vYFrom\| \leq \eta$, for a predefined $\eta > 0$. 

\textit{Closed-loop Prediction:} Given a state $\vX \in \vCXFree$, and an output trajectory $\vSigmaY \in \vClY$, the function $\fPropagate : (\vX,\vSigmaY) \mapsto \vSigmaX \in \vClX$ returns the state trajectory that is computed by simulating the system dynamics forward in time with the initial state $\vX$, and the reference trajectory $\vSigmaY$.

\textit{Collision Test:} Given two points $\vYFrom, \vYTo \in \vClGY$, the Boolean function $\PrcObstacleFree(\vYFrom, \vYTo)$ returns $\True$ if the line segment between $\vYFrom$ and $\vYTo$ lies in $\vCYFree$ 
and $\False$ otherwise.

\textit{Cost-to-come Values:} Given a graph $\vClGY = (\vCVY,\vCEY)$, let $\vCtocStarValue$ denote the optimal cost-to-come value of the node $\vVY \in \vCVY$ that can be achieved in $\vClGY$. Each node $\vVY \in \vCVY$ is associated with two estimates of the optimal cost-to-come value~(see~\cite{arslan2013useofrelaxation,koenig2004lifelongplanning}). The $g$-value of $\vVY$ is the cost of the path to $\vVY$ from a given initial state $\vYInit \in \vCYFree$. The one step look-ahead $g$-value of $\vVY$ is denoted with $\vCtocBarValue$ and defined as 
$$
 \vVY.\vCtocBarValue = 
  \begin{cases} 
      0,   & \text{if~}\vVY.\vY = \vYInit, \\
	\min\limits_{\vEY \in \vCEYPred} \left( \vVYPred.\vCtocValue + \fCostValue(\vSigma)\right), & \text{otherwise}, \\
  \end{cases}
$$
where $\vCEYPred = \PrcIncoming(\vClGY,\vVY)$, $\vVYPred = \vEY.\vTail$, and $\vSigma$ is the state trajectory that is computed via closed-loop prediction, i.e., the dynamical system is simulated forward in time with the initial state $\vVYPred.\vParentTrajectory.\fBack()$ and the reference trajectory $\vEY.\vSigma$. 

\textit{Heuristic Value:} Given a node $\vVY \in \vCVY$, and an output goal region $\vCYGoal$, the function $\fComputeHeuristic : (\vVY,\vCYGoal) \mapsto r$ returns an estimate $r$ of the optimal cost from $\vVY$ to $\vCYGoal$; it return zero if $\vVY \in \vCYGoal$. 
 In this paper, we always assume that $\fComputeHeuristic $ computes an admissible heuristic, that is, it never overestimates the actual cost of reaching $\vCYGoal$. 

\textit{Queue Operations:} Nodes of the computed graphs are associated with some keys and priority queues are used to sort these nodes based on the precedence relation between keys. The following functions are implemented to maintain a given priority queue $\vClQ$:

\begin{itemize}
\item $\vClQ.\PrcTopKey()$ returns the highest priority of all nodes in the priority queue $\vClQ$ with the smallest key value if the queue is not empty. If $\vClQ$ is empty, then $\vClQ.\PrcTopKey()$ returns a key value of $k = [\infty; \infty]$.


\item $\vClQ.\PrcPop()$ deletes the node with the highest priority in the priority queue $\vClQ$ and returns a reference to the node.

\item $\vClQ.\PrcUpdate(\vVY, k)$ sets the key value of the node $\vVY$ to $k$ and reorders the priority queue $\vClQ$.

\item $\vClQ.\PrcInsert(\vVY, k)$ inserts the node $\vVY$ into the priority queue $\vClQ$ with the key value $k$.

\item $\vClQ.\PrcRemove(\vVY)$ removes the node $\vVY$ from the priority queue $\vClQ$.
\end{itemize}

\textit{Initialization:} Given an initial point $\vXInit \in \vCX$, a goal region in the output space $\vCYGoal \subset \vCY$, the function $\PrcInitialize: (\vXInit, \vCYGoal) \mapsto (\vClGY,\vClGSigma,\vClQ, \vClQGoal)$ returns a graph $\vClGY$ that has only node $\vVY$, whose output point is $\vVY.\vY = \PrcOutputMap(\vXInit)$, a graph $\vClGSigma$ that has the only node $\vVSigma$, whose trajectory is a single point $\vVSigma.\vSigma = \vXInit$, and empty priority queues $\vClQ$ and $\vClQGoal$ that are used for ordering of nongoal and goal nodes, which represent points in $\vCY$, respectively.

\textit{Exploration:} Given a tuple of data structures $\vClS =(\vClGY,\vClGSigma,\vClQ, \vClQGoal)$, where $\vClGY$ and $\vClGSigma$ are graphs whose nodes represent points in $\vCY$ and trajectories in $\vClX$, respectively, and $\vClQ$ and $\vClQGoal$ are priority queues that are used for ordering of nongoal and goal nodes that represent points in $\vCY$, a goal region in the output space $\vCYGoal \subset \vCY$, and a point $\vY \in \vCY$, the function $\PrcExtend : (\vClS, \vCYGoal, \vY) \mapsto  \vClS^{\prime} = (\vClGY^{\prime},\vClGSigma^{\prime},\vClQ^{\prime}, \vClQGoal^{\prime})$ includes a new node, multiple edges to $\vClGY$ and multiple nodes, edges to $\vClGSigma$, updates the priorities of nodes in $\vClQ$ and $\vClQGoal$ and returns an updated tuple $\vClS^{\prime}$.

\textit{Exploitation:} Given a tuple of data structures $\vClS =(\vClGY,\vClGSigma,\vClQ, \vClQGoal)$, where $\vClGY$ and $\vClGSigma$ are graphs whose nodes represent points in $\vCY$ and trajectories in $\vClX$, respectively, and $\vClQ$ and $\vClQGoal$ are priority queues that are used for ordering of nongoal and goal nodes that represent points in $\vCY$, the function $\PrcReplan : \vClS \mapsto  \vClS^{\prime} = (\vClGY^{\prime},\vClGSigma^{\prime},\vClQ^{\prime}, \vClQGoal^{\prime})$ rewires the parent node of the nodes in $\vClGY$ based on their cost-to-come values, includes new nodes and edges in $\vClGSigma$ if necessary, that is, propagating dynamics of the system for new sequence of reference trajectories, and returns an updated tuple $\vClS^{\prime}$.

\textit{Construction of Solution:} Given a tuple of data structures $\vClS =(\vClGY,\vClGSigma,\vClQ, \vClQGoal)$,
the function $\PrcConstructSolution : \vClS \mapsto \vClTX$ returns a tree whose edges and nodes represent simulated trajectories in $\vClX$ and the corresponding internal states of the nodes of $\vClGY$. These trajectories are computed by propagating the dynamics with reference trajectories that are encoded in a tree of $\vClGY$, which is formed by the edges between nodes of $\vClGY$ and their parent nodes.

\textit{Graph and List Operations:} The following functions are used in the $\AlgClosedLoopRRTsharp$ algorithm.

\begin{itemize}
\item Given a node $v \in V$ in a directed graph $\graph{G}=(V,E)$, the set-valued function $\PrcSuccessor : (\graph{G}, v) \mapsto V^{\prime} \subseteq V$ returns the nodes in $V$ that are the heads of the edges emanating from $v$, that is, $
\PrcSuccessor(\graph{G}, v) :=  \left\{ v^{\prime} \in V: e.\vTail = v \text{~and~} e.\vHead = v^{\prime}, \, e \in E \right\}.$

\item Given a node $v \in V$ in a directed graph $\graph{G}=(V,E)$, the set-valued function $\PrcPredecessor : (\graph{G}, v) \mapsto V^{\prime} \subseteq V$ returns the nodes in $V$ that are the tails of the edges going into $v$, that is, $
\PrcPredecessor(\graph{G}, v) := \left\{ v^{\prime} \in V: e.\vTail = v^{\prime} \text{~and~} e.\vHead = v, \, e \in E \right\}.
$

\item Given a node $v \in V$ in a directed graph $\graph{G}=(V,E)$, the set-valued function $\PrcOutgoing : (\graph{G}, v) \mapsto E^{\prime} \subseteq E$ returns the edges in $E$ whose tail is $v$, that is, $\PrcOutgoing(\graph{G}, v) := \left\{ e \in E: e.\vTail = v \right\}.$

\item Given a node $v \in V$ in a directed graph $\graph{G}=(V,E)$, the set-valued function $\PrcIncoming : (\graph{G}, v) \mapsto E^{\prime} \subseteq E$ returns the edges in $E$ whose head is $v$, that is, $\PrcIncoming(\graph{G}, v) := \left\{ e \in E: e.\vHead = v \right\}.$
\end{itemize}


\begin{itemize}
\item Given a list of nodes $\vCVZ$, where its nodes represent points in $\vCZ$, and a point $\vZ \in \vCZ$, the function $\fFind : (\vCVZ,\vZ) \mapsto \vVZ \in \vCVZ$ returns the node in $\vCVZ$ that satisfies $\vVZ.\vZ = \vZ$ if there exists any such node, null  otherwise. 

\item Given a list of nodes $\vCVZ$, where its nodes represent points in $\vCZ$, the function $\fBack$ returns a reference to the last node in the list if it is not empty, and null  otherwise.

\item Given a list of nodes $\vCVZ$, where its nodes represent points in $\vCZ$, the function $\fFront$ returns a reference to the first node in the list if it is not empty, and null  otherwise.
\end{itemize}

\section{The \AlgClosedLoopRRTsharp Algorithm}\label{section:closed_loop_rrtsharp_algorithm}

\subsection{Details of Data Structures}
Each node $\vVY$ in the graph $\vClGY$ is an $\fInitOutputNode$ data structure, summarized in Table~\ref{table:reference_node_edge_structure}. Each node $\vVY$ is associated with a reference point $y \in \reals^{m}$. It contains two estimates of the optimal cost-to-come value between the initial reference point and $y$, namely, cost-to-come value $\vCtocValue$ and one step look-ahead $g$-value $\vCtocBarValue$. It also keeps a heuristic value $\vHValue$, which is an underestimate of the optimal cost value between $y$ and $\vCYGoal$, to guide and reduce the search effort. Whenever $\vCtocBarValue$ is updated during the replanning procedure, the reference node that yields the corresponding minimum cost-to-come value is stored in the parent reference node $\vParentReference$. Lastly, $\vParentTrajectory$ is the trajectory that is computed  by closed-loop prediction when the system is simulated with the reference trajectory between the nodes $\vParentReference$ and $\vVY$. Its terminal state represents the internal state associated with $\vVY$.

\begin{table}
\centering
\caption{{\small The node ($\PrcOutputNode$) and edge ($\PrcOutputEdge$) data structures for points and trajectories in output space, respectively}}\label{table:reference_node_edge_structure}
\begin{tabular}{c|l|p{5.1cm}}
{\bf field}    & {\bf type}	& {\bf description}       \\ 
\hline
\hline
$y$	&   vector $\in \mathbb{R}^{p}$     & output point associated with this node \\
$\vCtocValue$    &  real  $\in \mathbb{R}$        & cost-to-come value              \\
$\vCtocBarValue$    & real    $\in \mathbb{R}$       & one step look-ahead $g$-value         \\
$\vHValue$    &   real $\in \mathbb{R}$        & heuristic value for the cost between $y$ and $\mathcal{Y}_{\mathrm{goal}}$ \\
$\vParentReference$    & $\fInitOutputNode$ & reference to the parent output node        \\
$\vParentTrajectory$     & $\fInitTrajectoryNode$   & reference to the parent trajectory node       \\
\hline
$\referenceTrajectoryValue$          &   trajectory $\in \vClY$        	& output trajectory associated with this edge                       \\
$\tailRefNode$          &    $\fInitOutputNode$      	& reference to the tail output node         \\
$\headRefNode$          &  $\fInitOutputNode$          & reference to the head output node    \\
\hline
\end{tabular}
\end{table}

Each edge $\vEY$ in the graph $\vClGY$ is an $\fInitOutputEdge$ data structure, summarized in Table~\ref{table:reference_node_edge_structure}. Each edge $\vEY$ is associated with a trajectory  $r \in \vClY$. It also contains two output nodes, namely, $\vTail$ and $\vHead$, which represent the tail and the head output nodes of $\vEY$, respectively.

Each node $\vVSigma$ in the graph $\vClGSigma$ is a $\fInitTrajectoryNode$ data structure, summarized in Table~\ref{table:trajectory_node_edge_structure}. Each node $\vVSigma$ is associated with a trajectory $\sigma \in \vClX$. It contains an output edge $\vEY$, which corresponds to the reference trajectory that yields $\sigma$ as the closed-loop prediction. It also keeps a list of outgoing output edges $\vOutgoing$, and this list is used to compute outgoing trajectory nodes emanating from the terminal state of $\sigma$. 

\begin{table}
\centering
\caption{{\small The node ($\PrcTrajectoryNode$) and edge ($\PrcTrajectoryEdge$) data structures for trajectories in state space} }\label{table:trajectory_node_edge_structure}
\begin{tabular}{c|l|p{0.53\columnwidth}}
{\bf field} & {\bf type} & {\bf description}       \\ 
\hline
\hline
$\sigma$          &    trajectory $\in \vClX$     	& state trajectory associated with this node    \\
$e_{y}$           &   $\fInitOutputEdge$         & reference to the output edge 
\\
$\outgoingEdgeList$		   &   $\fInitOutputEdge$ array        & list of outgoing output edges   \\
\hline
$\sigma$          &    trajectory $\in \vClX$     	& state trajectory associated with this edge                      \\
$\tailRefNode$          &    $\fInitTrajectoryNode$     	&  reference to the tail trajectory node          \\
$\headRefNode$           &   $\fInitTrajectoryNode$        & reference to the head trajectory node     \\ 
\hline          
\end{tabular}

\end{table}

Each edge $\vESigma$ in the graph $\vClGSigma$ is a $\fInitTrajectoryEdge$ data structure, summarized in Table~\ref{table:trajectory_node_edge_structure}. Each edge $\vESigma$ is associated with a trajectory  $\sigma \in \vClX$. It contains two trajectory nodes, namely, $\vTail$ and $\vHead$ which represent the tail and the head trajectory nodes of $\vESigma$, respectively.   

\subsection{Details of the Procedures}

Algorithm~\ref{alg:body_closed_loop_rrtsharp} gives the body of the \AlgClosedLoopRRTsharp{} algorithm. First, the algorithm initializes the tuple of data structures $\vClS$ that is incrementally grown and updated as exploration and exploitation are performed (Line 3). The tuple $\vClS$ contains the graphs $\vClGY$ and $\vClGSigma$, which are used to store output nodes and state trajectory nodes, respectively, and the priority queues $\vClQ$ and $\vClQGoal$. The details of  $\PrcInitialize$  are given in Algorithm~\ref{alg:initialize_closed_loop_rrtsharp}.
The graph $\vClGSigma$ is created with no edges and $\vVSigma$ as its only node. This node represents a state trajectory that contains only the initial state $\vXInit$. Then, likewise, the graph $\vClGY$ is initialized with no edges and $\vVY$ as its only node that represents $\vYInit$. The $g$- and $\bar{g}$-values of $\vVY$ are set with zero cost value.
The parent trajectory node of $\vVY$ is set with the reference to the node $\vVSigma$.
{
\setlength{\intextsep}{1pt}
\IncMargin{0.7em}

\begin{algorithm}[h]

 	
 	\small
    \DontPrintSemicolon
    \SetKwInOut{Input}{input}
    \SetKwInOut{Output}{output}
    \SetKwBlock{NoBegin}{}{end}
	




    \SetKwData{vCVR}{$V_{r}$}
    \SetKwData{vEPrime}{$E^{\prime}$}
    \SetKwData{xinit}{$x_{\mathrm{init}}$}
    \SetKwData{vI}{$k$}
    \SetKwData{xRand}{$x_{\mathrm{rand}}$}
    \SetKwData{yrand}{$y_{\mathrm{rand}}$}
    \SetKwData{vRRand}{$r_{\mathrm{rand}}$}
    \SetKwData{vT}{$\mathcal{T}$}
    \SetKwData{vG}{$\mathcal{G}$}
    \SetKwData{XGoal}{$X_{\mathrm{goal}}$}
    \SetKwData{YGoal}{$Y_{\mathrm{goal}}$}
    \SetKwData{x}{$x$}
    

    \SetKwData{X}{$X$}
    
    \SetKwData{xParent}{$x_{\mathrm{parent}}$}
    \SetKwData{vCEX}{$E_{x}$}
	\SetKwData{vCVSigma}{$V_{\sigma}$}
    \SetKwData{vCESigma}{$E_{\sigma}$}

    \SetKwData{vxParent}{$v_{x,\mathrm{parent}}$}
	\SetKwData{cS}{$\mathcal{S}$}



    \SetFuncSty{textbf}
    \fRRTsharp{$\xinit,\XGoal,\X$}
    \SetFuncSty{texttt}
    \NoBegin
    {
	    $\YGoal \coloneqq \fOutputMap{\XGoal}$;
	    
		$\cS \leftarrow \fInitialize{\xinit,\YGoal}$;
		
        \For{$\vI = 1$ to $N$ \label{line:rrtsharp_itbegin}}
        {
            $\yrand \leftarrow \fSample{\vI}$;

            $\cS \leftarrow \fExtend{\cS,\YGoal,\yrand}$;

			\tikzmark{a0}$\S \leftarrow \fReplan{\cS}$;\tikzmark{a1}\label{line:rrtsharp_itend}
		}

		$\vClTX \leftarrow \fConstructSolution{\cS}$;
		
        \Return{$\vClTX$};
    }
    
\caption{{\small The $\AlgClosedLoopRRTsharp$ Algorithm}}\label{alg:body_closed_loop_rrtsharp}

\end{algorithm}
\DecMargin{2em}

}
{
	\setlength{\intextsep}{1pt}
	\IncMargin{0.7em}

\begin{algorithm}

 	
 	\small
    \DontPrintSemicolon
    \SetKwInOut{Input}{input}
    \SetKwInOut{Output}{output}
    \SetKwBlock{NoBegin}{}{end}
	




    \SetKwData{vy}{$v_{y}$}
    \SetKwData{Vy}{$V_{y}$}
    \SetKwData{Ey}{$E_{y}$}
        
    \SetKwData{Vsigma}{$V_{\sigma}$}
    \SetKwData{Esigma}{$E_{\sigma}$}    
    \SetKwData{vEPrime}{$E^{\prime}$}
    \SetKwData{xinit}{$x_{\mathrm{init}}$}
    \SetKwData{vI}{$k$}
    \SetKwData{vXRand}{$x_{\mathrm{rand}}$}
    \SetKwData{yrand}{$y_{\mathrm{rand}}$}
    \SetKwData{vT}{$\mathcal{T}$}
    \SetKwData{vG}{$\mathcal{G}$}
    \SetKwData{vClXGoal}{$\mathcal{X}_{\mathrm{goal}}$}
    \SetKwData{Ygoal}{$Y_{\mathrm{goal}}$}
    

    
    \SetKwData{vXParent}{$x_{\mathrm{parent}}$}
    \SetKwData{vCVX}{$V_{x}$}
    \SetKwData{vCEX}{$E_{x}$}
	\SetKwData{vCVSigma}{$V_{\sigma}$}
    \SetKwData{vCESigma}{$E_{\sigma}$}

    \SetKwData{vVX}{$v_{x}$}
    \SetKwData{vCVX}{$V_{x}$}

    \SetKwData{Vsigma}{$v_{\sigma}$}
    \SetKwData{psigma}{$\mathtt{p}_{\sigma}$}
    \SetKwData{vEX}{$e_{x}$}
	\SetKwData{Gy}{$\mathcal{G}_{y}$}
	\SetKwData{Gsigma}{$\mathcal{G}_{\sigma}$}
	\SetKwData{Tx}{$\mathcal{T}_{x}$}
	\SetKwData{Qgoal}{$\mathcal{Q}_{\mathrm{goal}}$}
	\SetKwData{vRInit}{$r_{\mathrm{init}}$}
	\SetKwData{yinit}{$y_{\mathrm{init}}$}
	\SetKwData{hvalue}{$\mathtt{h}$}
	\SetKwData{Sc}{$\mathcal{S}$}
	\SetKwData{Qc}{$\mathcal{Q}$}
	\SetKwData{ssigma}{$\sigma$}
	\SetKwData{vNull}{$\varnothing$}
	\SetKwData{gvalue}{$\mathtt{g}$}
	\SetKwData{gbarvalue}{$\mathtt{\bar{g}}$}
		


    \SetFuncSty{textbf}
    \fInitialize{$\xinit,\Ygoal$}
    \SetFuncSty{texttt}
    \NoBegin
    {
	    $\sigma \leftarrow \{\xinit\}$;
    		
    	$\Vsigma \leftarrow \fInitTrajectoryNode{\ssigma,\vNull,\vNull}$;
					    	  
		$\yinit \leftarrow \fOutputMap{\xinit}$;
		
		$\vy \leftarrow \fInitOutputNode{\yinit}$;
		
		$\vy.\gvalue \leftarrow 0$;
		$\vy.\gbarvalue  \leftarrow 0$;
		
		$\vy.\hvalue  \leftarrow \fComputeHeuristic{\yinit,\Ygoal}$;
		
		$\vy.\psigma \leftarrow \Vsigma$;
		
		$\Vy \leftarrow \{\vy\}$;
		$\Ey \leftarrow \varnothing$;
		
		$\vCVSigma \leftarrow \{\Vsigma\}$;
		$\vCESigma \leftarrow \varnothing$;
		
		$\Gy \leftarrow (\Vy,\Ey)$;
		$\Gsigma \leftarrow (\vCVSigma,\vCESigma)$;
				
		$\Qc \leftarrow \varnothing$;
		$\Qgoal \leftarrow \varnothing$;
		
		\Return{$\Sc \leftarrow (\Gy,\Gsigma,\Qc,\Qgoal)$};
	}
	
\caption{{\small The $\PrcInitialize$ Procedure}} \label{alg:initialize_closed_loop_rrtsharp}

\end{algorithm}
\DecMargin{0.7em}

}

The algorithm iteratively builds a graph of collision-free reference trajectories $\vClGY$ by first sampling an output point $\vYRand$ from the obstacle-free output space $\vCYFree$ (Line 5) and then extending the graph towards this sample (Line 6), at each iteration. The cost of the unique trajectory from the root node to a given node $\vVY$ is denoted as $\fCostValue(\vVY)$. It also builds another graph $\vClGSigma$, to store the state trajectories computed by simulation of the closed-loop dynamics when a reference trajectory is tracked. 
Once a new node is added to $\vClGY$ after  $\PrcExtend$,  $\PrcReplan$ is called to improve the existing solution by propagating the new information (Line 7). The dynamic system is simulated for different reference trajectories as needed during the search process. The computed state trajectories are added to the graph $\vClGSigma$ as new nodes along with the corresponding controls information. 

Finally, when a predetermined maximum number of iterations is reached,  $\PrcConstructSolution$ extracts the spanning tree of $\vClGY$ that contains the lowest-cost reference trajectories (Line 8). Algorithm~\ref{alg:construct_solution_closed_loop_rrtsharp} gives the details of $\PrcConstructSolution$.
{
	\setlength{\intextsep}{1pt}
	\IncMargin{0.7em}

\begin{algorithm}[h]

 	
 	\small
    \DontPrintSemicolon
    \SetKwInOut{Input}{input}
    \SetKwInOut{Output}{output}
    \SetKwBlock{NoBegin}{}{end}
	




    \SetKwData{vy}{$v_{y}$}
    \SetKwData{Vy}{$V_{y}$}
    \SetKwData{Ey}{$E_{y}$}
    \SetKwData{vI}{$k$}
    \SetKwData{vT}{$\mathcal{T}$}
    

    \SetKwData{vcapx}{$X$}
    
    \SetKwData{vXParent}{$x_{\mathrm{parent}}$}
    \SetKwData{vCVX}{$V_{x}$}
    \SetKwData{vCEX}{$E_{x}$}
    
    \SetKwData{vVXParent}{$v_{x,\mathrm{parent}}$}
    \SetKwData{vVX}{$v_{x}$}
    \SetKwData{vCVX}{$V_{x}$}

    \SetKwData{psigma}{$\mathtt{p}_{\sigma}$}
    \SetKwData{vEX}{$e_{x}$}
	\SetKwData{Gy}{$\mathcal{G}_{y}$}
	\SetKwData{Gsigma}{$\mathcal{G}_{\sigma}$}
	\SetKwData{vFrontier}{$\mathcal{Q}$}
	\SetKwData{Tx}{$\mathcal{T}_{x}$}
	\SetKwData{Qgoal}{$\mathcal{Q}_{\mathrm{goal}}$}
	\SetKwData{Sc}{$\mathcal{S}$}
	\SetKwData{sigmav}{$\sigma$}


    \SetFuncSty{textbf}
    
    \fConstructSolution{$\Sc$}
    \SetFuncSty{texttt}
    \NoBegin
    {
	    $(\Gy,\Gsigma,\vFrontier,\Qgoal) \leftarrow \Sc$;
	    
        $(\Vy,\Ey) \leftarrow \Gy$;
		$\vcapx \leftarrow \varnothing$;
		
        \ForEach{$\vy \in \Vy$}
        {

	        $\sigma \leftarrow \vy.\psigma.\sigmav$;

	        $\vVX \leftarrow \fInitStateNode{\sigmav.\fBack{}}$;
	        
	        $\vCVX \leftarrow \vCVX \cup \{\vVX\}$;
	        
	        $\vVXParent \leftarrow \fFind{\vCVX,\sigmav.\fFront{}}$;
	        
	        \If{$\vVXParent = \varnothing$}
  	        {
  		        $\vVXParent \leftarrow \fInitStateNode{\sigmav.\fFront{}}$;
  		        
  		        $\vCVX \leftarrow \vCVX \cup \{\vVXParent\}$;
  	        }
  	        
   	        $\vEX \leftarrow \fInitStateEdge{\vVXParent, \vVX, \sigmav}$;
   	        
            $\vCEX \leftarrow \vCEX \cup \{\vEX\}$;
              
            $\vcapx \leftarrow \vcapx \cup \{\sigma.\fBack{}\}$;	        
	        	        
        }

        \Return{$\Tx = (\vCVX,\vCEX)$};
    }

\caption{{\small The $\PrcConstructSolution$ Solution Procedure}}
\label{alg:construct_solution_closed_loop_rrtsharp}
\end{algorithm}
\DecMargin{0.7em}

}

\subsubsection{The $\PrcExtend$ Procedure}
\begin{figure}
	\centering
	\includegraphics[scale=0.3]{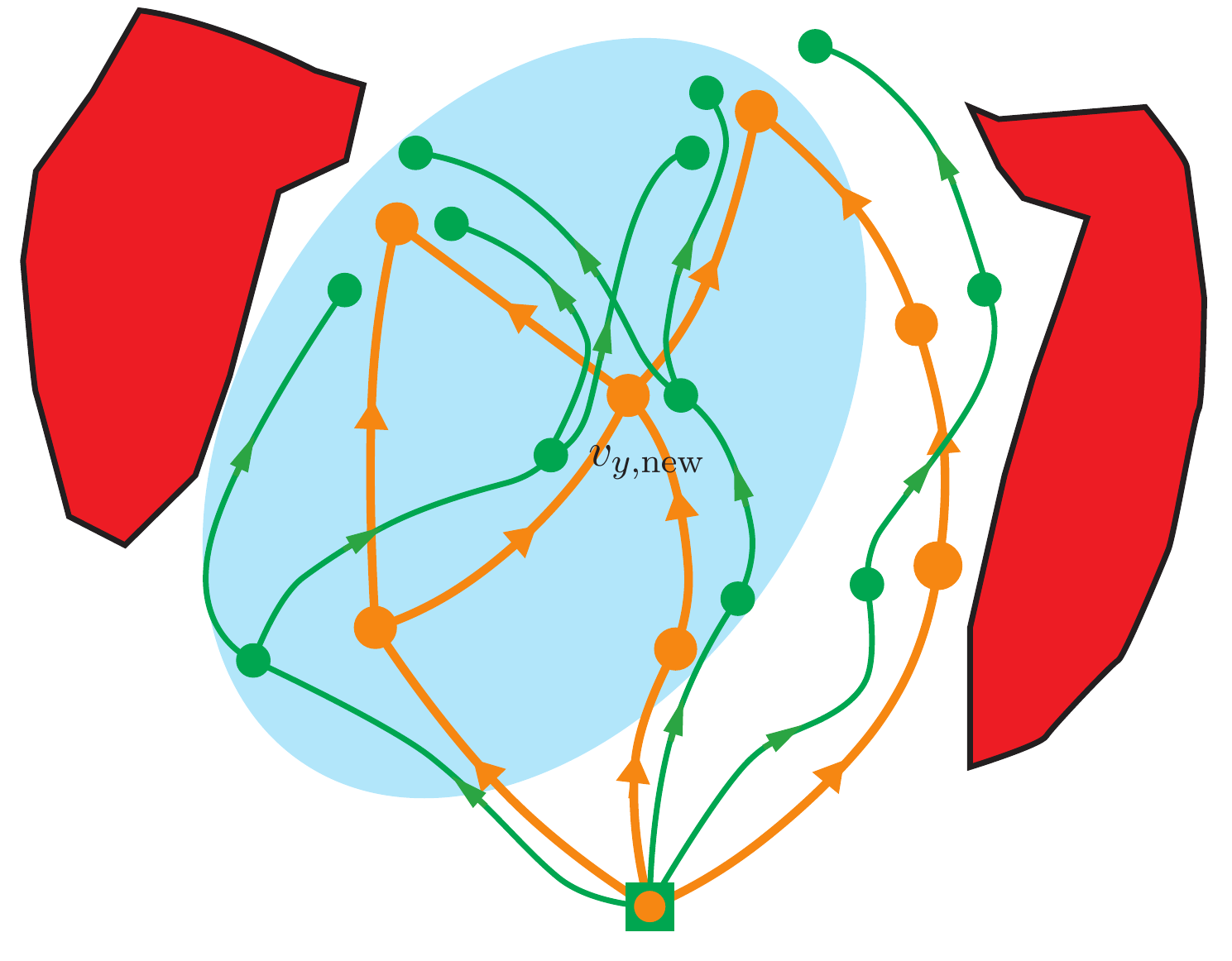}
	
	\caption{Extension of the graphs computed by the \AlgClosedLoopRRTsharp{} algorithm. Trajectories in the output and state spaces are shown in orange and green colors, respectively. Whenever a new node in the output space is added, then several incoming and outgoing edges are included to the graph in the vicinity of the new node, i.e., region colored with cyan.}\label{figure:cl_rrtsharp_extension}
	\vspace*{-15pt}
\end{figure}

The $\PrcExtend$ procedure is given in Algorithm~\ref{alg:extend_closed_loop_rrtsharp}. It first extends the nearest output node $\vVYNearest$ to the output sample $\vY$ (Lines 4-5). The output trajectory that extends the nearest output node $\vVYNearest$ towards the output sample $\vY$ is denoted as $\vRNew$. The final output point on the output trajectory $\vRNew$ is denoted as $\vYNew$. If $\vRNew$ is collision-free, then a new output node $\vVYNew$ is created to represent the new output point $\vYNew$ (Line 8), and the following changes in the vicinity of $\vVYNew$ on both graphs are shown in Fig.~\ref{figure:cl_rrtsharp_extension}. The initial node is shown as a square box, the obstacles are shown in red color, and the graphs $\vClGY$ and $\vClGSigma$ are  shown in orange and green colors.

{
	\setlength{\intextsep}{1pt}
	\IncMargin{0.7em}

\begin{algorithm}[h]
    \small

    \DontPrintSemicolon
    \SetKwInOut{Input}{input}
    \SetKwInOut{Output}{output}
    \SetKwBlock{NoBegin}{}{end}





    \SetKwData{Ygoal}{$Y_{\mathrm{goal}}$}
    \SetKwData{Xgoal}{$X_{\mathrm{goal}}$}
    \SetKwData{Tx}{$\mathcal{T}_{x}$}
    \SetKwData{Ty}{$\mathcal{T}_{y}$}
    \SetKwData{Gx}{$\mathcal{G}_{x}$}
    \SetKwData{Gy}{$\mathcal{G}_{y}$}
    \SetKwData{vTPrime}{$\mathcal{T}^{\prime}$}
    \SetKwData{vGPrime}{$\mathcal{G}^{\prime}$}
    \SetKwData{xalg}{$x$}
    \SetKwData{xgoal}{$x_{\mathrm{goal}}$}
    \SetKwData{vGRX}{$V_{x}$}
    \SetKwData{vEX}{$E_{x}$}
    \SetKwData{Vy}{$V_{y}$}
    \SetKwData{Ey}{$E_{y}$}
    
    \SetKwData{vFrontier}{$\mathcal{Q}$}
    \SetKwData{Qgoal}{$\mathcal{Q}_{\mathrm{goal}}$}
    \SetKwData{Ygoal}{$Y_{\mathrm{goal}}$}
           
    \SetKwData{vESigmaPrime}{$E_{\sigma}^{\prime}$}    
    \SetKwData{vXNearest}{$x_{\mathrm{nearest}}$}
    \SetKwData{vXNew}{$x_{\mathrm{new}}$}
    \SetKwData{vCXNear}{$\mathcal{X}_{\mathrm{near}}$}
    \SetKwData{vXNear}{$x_{\mathrm{near}}$}
	\SetKwData{xpred}{$x_{\mathrm{pred}}$}

	\SetKwData{vSigmaNear}{$\sigma_{\mathrm{near}}$}
	\SetKwData{ynew}{$y_{\mathrm{new}}$}
	\SetKwData{vYNewPrime}{$y^{\prime}_{\mathrm{new}}$}
		
	\SetKwData{yalg}{$y$}
	\SetKwData{vCYNear}{$\mathcal{R}_{\mathrm{near}}$}
    \SetKwData{vYNearest}{$r_{\mathrm{nearest}}$}
    \SetKwData{vYNear}{$r_{\mathrm{near}}$}
    \SetKwData{Gsigma}{$\mathcal{G}_{\sigma}$}
    
    \SetKwData{Gy}{$\mathcal{G}_{y}$}
    
    \SetKwData{vypred}{$v_{y,\mathrm{pred}}$}
    \SetKwData{vCVPred2}{$V_{\mathrm{pred}}$}
    \SetKwData{vCEPred}{$E_{y,\mathrm{pred}}$}

    \SetKwData{vCVSucc}{$V_{\mathrm{succ}}$}
    \SetKwData{vCESucc}{$E_{y,\mathrm{succ}}$}
        
	\SetKwData{vV2}{$y$}
	\SetKwData{vCVNear2}{$V_{\mathrm{near}}$}
	\SetKwData{vynearest}{$v_{y,\mathrm{nearest}}$}
	\SetKwData{vynew}{$v_{y,\mathrm{new}}$}
	\SetKwData{vynear}{$v_{y,\mathrm{near}}$}
    \SetKwData{psigma}{$\mathtt{p}_{\sigma}$}
     
    \SetKwData{ey}{$e_{y}$}
    \SetKwData{esigma}{$e_{\sigma}$}
    
    \SetKwData{vCVSigmaPrime}{$V^{\prime}_{\sigma}$}
    \SetKwData{vCESigmaPrime}{$E^{\prime}_{\sigma}$}
    
    \SetKwData{vCVSigma}{$V_{\sigma}$}
    \SetKwData{vCESigma}{$E_{\sigma}$}
    
    \SetKwData{vsigmapred}{$v_{\sigma,\mathrm{pred}}$}
    \SetKwData{vsigmanear}{$v_{\sigma,\mathrm{near}}$}
    \SetKwData{vsigmanew}{$v_{\sigma,\mathrm{new}}$}
    \SetKwData{rnew}{$r_{\mathrm{new}}$}
	
	\SetKwData{bargvalue}{$\mathtt{\bar{g}}$}
	\SetKwData{pyref}{$\mathtt{p}_{y}$}
	\SetKwData{psigma}{$\mathtt{p}_{\sigma}$}
	\SetKwData{vCostValue}{$\mathtt{c}$} 
	
	\SetKwData{tthead}{$\mathtt{head}$}
	\SetKwData{ralg}{$r$}
	\SetKwData{gvalc}{$\mathtt{g}$}
	\SetKwData{hvalueext}{$\mathtt{h}$}
	\SetKwData{Scval}{$\mathcal{S}$}
	\SetKwData{sigmaalg}{$\sigma$}


    \SetFuncSty{textbf}
    \fExtend{$\Scval,\Xgoal,\yalg$}
    \SetFuncSty{texttt}
    \NoBegin
    {
		$(\Gy,\Gsigma,\vFrontier,\Qgoal) \leftarrow \Scval$;
		
        $(\Vy,\Ey) \leftarrow \Gy$;
        $(\vCVSigma,\vCESigma) \leftarrow \Gsigma$;
        
        $\vynearest \leftarrow \fNearest{\Gy,\yalg}$;

        $\rnew \leftarrow \fSteer{\vynearest.\yalg,\yalg}$;

        \If{$\fObstacleFree{\rnew}$}
        {
            $\ynew \leftarrow \rnew.\fBack{}$;
            
            $\vynew \leftarrow \fInitOutputNode{\ynew}$;
            
            $\vynew.\hvalueext \leftarrow \fComputeHeuristic{\ynew,\Ygoal}$;
            
            $\vCVNear2 \leftarrow \fNear{\Gy,\ynew,\ensuremath{|\Vy|}} \cup \{ \vynearest \}$;

			$\vCESucc \leftarrow \varnothing$;
			$\vCEPred \leftarrow \varnothing$;
			
			\ForEach{$\vynear \in \vCVNear2$}
            {
	            $\ralg \leftarrow \fSteer{\ynew,\vynear.\yalg}$;
	            
                \If{$\fObstacleFree{\ralg}$}
                {
	                $\ey \leftarrow \fInitOutputEdge{\vynew,\vynear,\ralg}$;
	                
	                $\vCESucc \leftarrow \vCESucc \cup \{ \ey \}$;
                }
                
	            $\ralg \leftarrow \fSteer{\vynear.\yalg,\ynew}$;
	           
                \If{$\fObstacleFree{\ralg}$}
                {
	                $\ey \leftarrow \fInitOutputEdge{\vynear,\vynew,\ralg}$;
	               
	                $\vCEPred \leftarrow \vCEPred \cup \{ \ey \}$;
                }
            }
            
            $\vCVSigmaPrime \leftarrow \varnothing$;
            $\vCESigmaPrime \leftarrow \varnothing$;
            
            \ForEach{$\ey \in \vCEPred$}
            {
	            $\vypred \leftarrow \ey.\vTail$;
	            
	            $\vsigmapred \leftarrow \vypred.\psigma$;
	              
	            $\xpred \leftarrow \vsigmapred.\sigma.\fBack{}$; 
	              
	            $\sigma \leftarrow \fPropagate{\xpred,\ey.\ralg}$;
	             
	            \If{$\fObstacleFree{\sigmaalg}$} 
	            { 
		            $\vsigmanew \leftarrow \fInitTrajectoryNode{\sigmaalg,\ey,\vCESucc}$;
	             
		            $\esigma \leftarrow \fInitTrajectoryEdge{\vsigmapred,\vsigmanew,\sigmaalg}$;
	            
		            $\vCVSigmaPrime \leftarrow \vCVSigmaPrime \cup \{ \vsigmanew \}$;
		            
		            $\vCESigmaPrime \leftarrow \vCESigmaPrime\cup \{ \esigma \}$;
	                  
		            \If{$\vynew.\bargvalue >  \vypred.\gvalc + \fCostValue{\sigmaalg}$}
		            {
			         	$\vynew.\bargvalue \leftarrow  \vypred.\gvalc + \fCostValue{\sigmaalg}$;
	                      
		                $\vynew.\pyref \leftarrow \vypred$;
	                  
		                $\vynew.\psigma \leftarrow \vsigmanew$;
	
		            }
	
                }
	         } 

            $\Vy \leftarrow \Vy \cup \{ \vynew \}$;
            $\Ey \leftarrow \Ey \cup \vCESucc \cup \vCEPred$;
            
            $\vCVSigma \leftarrow \vCVSigma \cup \vCVSigmaPrime$;
            $\vCESigma \leftarrow \vCESigma \cup \vCESigmaPrime$;

            $\Gy \leftarrow (\Vy,\Ey)$;
            $\Gsigma \leftarrow (\vCVSigma,\vCESigma)$;
            
            $\vFrontier \leftarrow \fUpdateQueue{\vFrontier,\vynew}$;
                                
            $\Qgoal \leftarrow \fUpdateGoal{\Qgoal,\vynew,\Xgoal}$;
                        
        }
        
        \Return{$\Scval \leftarrow (\Gy,\Gsigma,\vFrontier,\Qgoal)$};
    }

\caption{{\small The ${\tt Extend}$ Procedure}{\color{white}$^\#$}}\label{alg:extend_closed_loop_rrtsharp}

\end{algorithm}
\DecMargin{0.7em} 
}

 The members of the node $\vVYNew$ are set as follows. First,  $\PrcNear$ finds the set of neighbor output nodes $\vCVNear$ in the neighborhood of the new output point $\vYNew$ (Line 9). Then, the set of incoming edges $\vCEYPred$ and outgoing edges $\vCEYSucc$ of the new output node $\vVYNew$ are computed by using the information of the neighbor output nodes (Lines 10-19). 
 
 Once the new output node $\vVYNew$ is created together with the set of incoming edges $\vCEYPred$ and outgoing edges $\vCEYSucc$ connecting it to its neighbor output nodes $\vCVNear$,  $\PrcExtend$  attempts to find the best incoming edge that yields a segment of a reference trajectory which incurs minimum cost to get to $\vVYNew$ among all incoming edges in $\vCEYPred$ (Lines 20-34). That is, for any incoming edge $\vEY$ in $\vCEYPred$, the algorithm first gets the information of the predecessor output node $\vVYPred$ and its internal state $\vXPred$ by using the information of the parent state trajectory node $\vVSigmaPred$ (Lines 22-24). Then, it simulates the system  forward in time with the state $\vXPred$ being the initial state and $\vEY.\vR$ being the reference trajectory to be tracked, (Line 25). If the state trajectory $\vSigma$ computed by closed-loop prediction is collision-free, a new trajectory node $\vVSigmaNew$ is created together with its list of outgoing output trajectories being initialized with $\vCEYSucc$ (Line 27). When a new trajectory node $\vVSigmaNew$ is created, the outgoing state trajectories emanating from the final state of the state trajectory $\vVSigmaNew.\vSigma$ via closed-loop prediction are not immediately computed, for the sake of efficiency. Instead, the algorithm keeps the set of candidate outgoing output trajectories, that is, the edges in $\vCEYSucc$, in a list $\vVSigmaNew.\vOutgoing$, and the simulation of the system  for these output trajectories is postponed until the head output node of the output edge $\vVSigmaNew.\vEY$ is selected for the Bellman update during the $\PrcReplan$ procedure. Once the new state trajectory node $\vVSigmaNew$ and the edge between the predecessor state trajectory node $\vVSigmaPred$ and itself are created (Lines 27-28), they are added to the set of nodes and edges of the graph $\vClGSigma$, respectively (Lines 29-30). If the incoming output edge $\vEY$ between the predecessor output node $\vVYPred$ and the new output node $\vVYNew$ yields a collision-free state trajectory $\vSigma$ that incurs cost less than the current cost of $\vVYNew$, then, the $\vCtocBarValue$-value of $\vVYNew$ is set with new lower cost, $\vVYPred$ and $\vVSigmaNew$ are made the new parent output node and the new parent state trajectory node of $\vVYNew$ (Lines 31-34). 
 
 After successful creation of the new output node $\vVYNew$, it is added to the graph $\vClGY$ together with all of its collision-free output edges (Line 36). Likewise, all trajectory nodes and edges created during the simulation of the system dynamics are added to the graph $\vClGSigma$ (Line 37). Lastly, the priority queues, $\vClQ$ and $\vClQGoal$ are updated accordingly by using the information of the new output node $\vVYNew$, that is, reordering of the priorities after insertion of $\vVYNew$ to the queue $\vClQ$ and reordering the goal output nodes in $\vClQGoal$ if $\vVYNew$ happens to be a goal output node (Lines 38-39).
\subsubsection{The $\PrcReplan$ Procedure}
The $\PrcReplan$ procedure is given in Algorithm~\ref{alg:replan_closed_loop_rrtsharp} (see~\cite{arslan2013useofrelaxation}). It improves cost-to-come values of output nodes by operating on the nonstationary and promising nodes of the graph $\vClGY$. It  pops the most promising nonstationary node from the priority queue $\vClQ$, if there are any, and this  node is made stationary by assigning its $\vCtocBarValue$-value to its $\vCtocValue$-value (Lines 5-6). Then, the $\vCtocValue$-value of the output node $\vVY$ is used to improve the $\vCtocBarValue$-values of its neighbor output nodes. Before this, the algorithm computes the set of all outgoing state trajectories emanating from internal state of the output node $\vV$ (Lines 9-16). To do so, the algorithm first gets the information of the internal state $\vX$ by using the parent state trajectory node of $\vVY$ (Lines 7-8). For any outgoing edge $\vEY$ in $\vVSigma.\vOutgoing$, the algorithm first gets the information of the successor output node $\vVYSucc$ by using the output edge $\vEY$ (Line 10). Then, it simulates the system  forward in time with the state $\vX$ being the initial state and $\vEY.\vR$ being the reference trajectory to be tracked (Line 11). If the state trajectory $\vSigma$ computed by closed-loop prediction is collision-free, a new trajectory node $\vVSigmaSucc$ is created together with its list of outgoing output trajectories being initialized with the set of outgoing output edges of $\vVYSucc$ (Line 13). Also, a state trajectory edge between $\vVSigma$ and $\vVSigmaSucc$ is created (Line 14). Then, the new state trajectory node and edge are tentatively added to the set of nodes and edges of the graph $\vClGSigma$ (Lines 15-16). This  continues until all candidate outgoing output trajectories are processed in the closed-loop simulation, then the list $\vVY.\vOutgoing$ is cleared up (Line 17). All newly computed state trajectory nodes and edges are added to the graph $\vClGSigma$ (Line 18).

 For each outgoing state trajectory $\vSigma$, $\PrcReplan$ adds up its cost, incurred by reaching to the successor output node $\vVYSucc$ to the $\vCtocValue$-value of $\vVY$, and compare it with the current $\vCtocBarValue$-value of $\vVYSucc$ (Line 22). If the outgoing state trajectory edge $\vSigma$ yields a lower cost than $\vVYSucc$,  the $\vCtocBarValue$-value of $\vVYSucc$ is set with new lower cost, and $\vVY$ and $\vVSigmaSucc$ are made the new parent output node and the new parent state trajectory node of $\vVYSucc$, respectively (Lines 23-25). Last, the priority queues $\vClQ$ and $\vClQGoal$ are updated  by using the update information of the successor output node $\vVYSucc$, that is, reordering of the priorities after updating the key value of $\vVYSucc$ to the queue $\vClQ$ and reordering the goal output nodes in $\vClQGoal$ if $\vVYSucc$ happens to be a goal output node (Lines 26-27). These steps are repeated until there is no promising nonstationary output node left in the priority queue $\vClQ$, that is,  $\vClQ.\fTopKey() \succeq \vClQGoal.\fTopKey()$.
{
\setlength{\intextsep}{1pt}
\IncMargin{0.7em}

\begin{algorithm}[h]

    \small
    \DontPrintSemicolon
    \SetKwInOut{Input}{input}
    \SetKwInOut{Output}{output}
    \SetKwBlock{NoBegin}{}{end}





    \SetKwData{Gsigma}{$\mathcal{G}_{\sigma}$}
    \SetKwData{Gy}{$\mathcal{G}_{y}$}
    
    \SetKwData{Valg}{$V$}
    \SetKwData{vFrontier}{$\mathcal{Q}$}
    \SetKwData{Qgoal}{$\mathcal{Q}_{\mathrm{goal}}$}
    \SetKwData{Ygoal}{$Y_{\mathrm{goal}}$}
    \SetKwData{xgoal}{$x_{\mathrm{goal}}$}
    \SetKwData{vT}{$\mathcal{T}$}
    \SetKwData{vG}{$\mathcal{G}$}
    \SetKwData{vTPrime}{$\mathcal{T}^{\prime}$}
    \SetKwData{vGPrime}{$\mathcal{G}^{\prime}$}
    \SetKwData{Xgoal}{$X_{\mathrm{goal}}$}
    \SetKwData{vXminGoal}{$v^{*}_{\mathrm{goal}}$}
    \SetKwData{vVminGoal}{$v^{*}_{\mathrm{goal}}$}
    \SetKwData{xalg}{$x$}

	\SetKwData{pyalg}{$\mathtt{p}_{y}$}
	\SetKwData{psigma}{$\mathtt{p}_{\sigma}$}
	
	\SetKwData{vysucc}{$v_{y,\mathrm{succ}}$}
	\SetKwData{vy}{$v_{y}$}
	\SetKwData{bargvalue}{$\mathtt{\bar{g}}$}
	\SetKwData{gvalue}{$\mathtt{g}$}
	\SetKwData{vCostValue}{$\mathtt{c}$} 
	\SetKwData{ralg}{$r$} 
	\SetKwData{sigmaalg}{$\sigma$} 
		
	\SetKwData{ey}{$e_{y}$}
	\SetKwData{outgoingalg}{$\mathtt{outgoing}$} 
	\SetKwData{vsigma}{$v_{\sigma}$} 
	\SetKwData{vsigmasucc}{$v_{\sigma,\mathrm{succ}}$} 
	\SetKwData{esigma}{$e_{\sigma}$} 
    \SetKwData{Vsigma}{$V_{\sigma}$}
    \SetKwData{Esigma}{$E_{\sigma}$}   
    \SetKwData{headalg}{$\mathtt{head}$}  
	\SetKwData{vVYFrom}{$v_{y,\mathrm{from}}$} 
	\SetKwData{vVYTo}{$v_{y,\mathrm{to}}$} 
	\SetKwData{vVMinRGoal}{$v^{*}_{r,\mathrm{goal}}$} 
	\SetKwData{vCVYGoal}{$V_{y,\mathrm{goal}}$} 
	\SetKwData{vCVGoal}{$V_{\mathrm{goal}}$} 
	\SetKwData{vKey}{$\mathtt{key}$}  
	\SetKwData{vVSigmaPrime}{$v^{\prime}_{\sigma}$} 
	\SetKwData{vVRPrime}{$v^{\prime}_{r}$} 
	\SetKwData{Sc}{$\mathcal{S}$}
	

    \SetFuncSty{textbf}
    \fReplan{$\Sc, \Xgoal$}
    \SetFuncSty{texttt}
    \NoBegin
    {
	    $(\Gy,\Gsigma,\vFrontier,\Qgoal) \leftarrow \Sc$;

        $(\Vsigma,\Esigma) \leftarrow \Gsigma$;

        \tikzmark{p6st}\While{$\vFrontier.\fTopKey{} \prec \Qgoal.\fTopKey{}$}
        {
			$\vy \leftarrow \vFrontier.\fPop{}$;
				          
			$\vy.\gvalue\leftarrow \vy.\bargvalue$; 
				            
			$\vsigma \leftarrow \vy.\psigma$;
			              
			$\xalg \leftarrow \vsigma.\vSigma.\fBack{}$;
			              
			\ForEach{$\ey \in \vsigma.\outgoingalg$}
			{
				$\vysucc \leftarrow \ey.\headalg$;
			
				$\sigmaalg \leftarrow \fPropagate{\xalg,\ey.\ralg}$;
			
				\If{$\fObstacleFree{\sigmaalg}$}
				{
					$\vsigmasucc \leftarrow \fInitTrajectoryNode{\sigmaalg,\ey,\fOutgoing{\Gy,\vysucc}}$;

					$\esigma \leftarrow \fInitTrajectoryEdge{\vsigma,\vsigmasucc,\sigmaalg}$;
			
					$\Vsigma \leftarrow \Vsigma \cup \{ \vsigmasucc \}$;
				  
					$\Esigma \leftarrow \Esigma \cup \{ \esigma \}$;
				}
			}	
			              
			$\vsigma.\outgoingalg \leftarrow \varnothing$;
			
			$\Gsigma \leftarrow (\Vsigma,\Esigma)$;
				  		
			\tikzmark{p8st}\ForEach{$\vsigmasucc \in \fSuccessor{\Gsigma,\vsigma}$}
			{
				$\sigmaalg \leftarrow \vsigmasucc.\sigmaalg$;
			
				$\vysucc \leftarrow \vsigmasucc.\ey.\headalg$;
			
				\If{$\vysucc.\bargvalue > \vy.\gvalue + \fCostValue{\sigmaalg}$}
				{
			
					\tikzmark{p9st}$\vysucc.\bargvalue \leftarrow \vy.\gvalue + \fCostValue{\sigmaalg}$; 
			
					$\vysucc.\pyalg \leftarrow \vy$;
			
					$\vysucc.\psigma \leftarrow \vsigmasucc$;
						
					$\vFrontier \leftarrow \fUpdateQueue{\vFrontier,\vysucc}$;
			
					$\Qgoal \leftarrow \fUpdateGoal{\Qgoal,\vysucc,\Xgoal}$;\tikzmark{p6en}
			
				}
			}			            	
			            				       
        }

        \Return{$\Sc \leftarrow (\Gy,\Gsigma, \vFrontier, \Qgoal)$};
   }

\caption{ {\small $\PrcReplan$ Procedure}{\color{white}$^\#$}} \label{alg:replan_closed_loop_rrtsharp}

\end{algorithm}
\DecMargin{0.7em} 
}

The auxiliary procedures  in  $\PrcExtend$ and $\PrcReplan$  are shown in Algorithm~\ref{alg:auxiliary_procedures_closed_loop_rrtsharp}.  $\fUpdateQueue$  maintains the priority queue $\vClQ$ whenever a new output node is created or key value of an output node that is already in the queue is updated. During a call to  $\fUpdateQueue$  with the priority queue $\vClQ$ and the output node $\vVY$, there are three possible cases. First, if $\vVY$ is a nonstationary output node, that is, $\vVY.\vCtocValue \neq \vVY.\vCtocBarValue$, key value of $\vVY$ is updated and priorities in the queue are reordered  (Line 3). Second, if $\vVY$ is a nonstationary output node and it is not in the queue, then it is inserted to the queue $\vClQ$ with its key value (Line 5). Third, if $\vVY$  is a stationary output node, that is, $\vVY.\vCtocValue = \vVY.\vCtocBarValue$, and it is in the queue $\vClQ$, then, it is removed from the queue $\vClQ$ (Line 7).
{
\setlength{\intextsep}{1pt}
\IncMargin{0.7em}

\begin{algorithm}[h]
    \small
    \DontPrintSemicolon
    \SetKwInOut{Input}{input}
    \SetKwInOut{Output}{output}
    \SetKwBlock{NoBegin}{}{end}





    \SetKwData{Valg}{$V$}
    \SetKwData{vT}{$\mathcal{T}$}
    \SetKwData{vG}{$\mathcal{G}$}
    \SetKwData{vTPrime}{$\mathcal{T}^{\prime}$}
    \SetKwData{vGPrime}{$\mathcal{G}^{\prime}$}
    \SetKwData{vClXGoal}{$\mathcal{X}_{\mathrm{goal}}$}
    \SetKwData{Xgoal}{$X_{\mathrm{goal}}$}
    \SetKwData{Ygoal}{$Y_{\mathrm{goal}}$}
    \SetKwData{xalg}{$x$}
    \SetKwData{vK}{$k$}
    
	\SetKwData{yalg}{$y$}
	\SetKwData{sigmaalg}{$\sigma$}
    \SetKwData{gvalue}{$\mathtt{g}$}
	\SetKwData{gbarvalue}{$\mathtt{\bar{g}}$}
	\SetKwData{hvalue}{$\mathtt{h}$}
	\SetKwData{vFrontier}{$\mathcal{Q}$}
	\SetKwData{Qgoal}{$\mathcal{Q}_{\mathrm{goal}}$}
	\SetKwData{Ygoal}{$\mathcal{Y}_{\mathrm{goal}}$}
	\SetKwData{vyalg}{$v_{y}$}
	\SetKwData{vsigma}{$v_{\sigma}$}


    \SetFuncSty{textbf}
    \fUpdateQueue{$\vFrontier,\vyalg$}
    \SetFuncSty{texttt}
    \NoBegin
    {
         \If{$\vyalg.\gvalue  \neq \vyalg.\gbarvalue \text{~and~} \vyalg \in \vFrontier$}
         {
            $\vFrontier.\fUpdate{\vyalg,\fKey{\vyalg}}$;
         }
         \ElseIf{$\vyalg.\gvalue  \neq \vyalg.\gbarvalue \text{~and~} \vyalg \notin \vFrontier$}
         {
            $\vFrontier.\fInsert{\vyalg,\fKey{\vyalg}}$;
         }
         \ElseIf{$\vyalg.\gvalue  = \vyalg.\gbarvalue \text{~and~} \vyalg \in \vFrontier$}
         {
            $\vFrontier.\fRemove{\vyalg}$;
         }
         
         \Return{$\vFrontier$};
    }

	\SetFuncSty{textbf}
    \fUpdateGoal{$\Qgoal,\vyalg,\Ygoal$}
    \SetFuncSty{texttt}
    \NoBegin
    {
	    $\vsigma \leftarrow \vyalg.\vParentTrajectory$;
	    
	    $\xalg \leftarrow \vsigma.\sigmaalg.\fBack{}$;
	    
	    \If{$\xalg \in \Xgoal$}
	    {
		    \If{$\vyalg \in \Qgoal$}
	        {
		        $\Qgoal.\fUpdate{\vyalg,\fKey{\vyalg}}$;
	         }
	        \Else
	        {
		        $\Qgoal.\fInsert{\vyalg,\fKey{\vyalg}}$;
	        }
	    }
        \Return{$\Qgoal$};
    }
    
    \SetKwFunction{fH}{h}

    \SetKwData{vGPrime}{$g^{\prime}$}
    \SetKwData{vGMin}{$g_{\min}$}
    \SetKwData{vF}{$f$}
    \SetKwData{vKey}{$key$}

    \SetFuncSty{textbf}
    \fKey{$\vyalg$}
    \SetFuncSty{texttt}
    \NoBegin
    {
        \Return{$\vK = (\vyalg.\gbarvalue+ \vyalg.\hvalue, \vyalg.\hvalue)$};
    }

\caption{{\small Auxiliary Procedures}{\color{white}$^\#$}} \label{alg:auxiliary_procedures_closed_loop_rrtsharp}

\end{algorithm}
\DecMargin{0.7em} 
}

Algorithm~\ref{alg:constructor_procedures_closed_loop_rrtsharp} gives constructor procedures for node and edge data structures used in the $\AlgClosedLoopRRTsharp$.

{
\setlength{\intextsep}{1pt}

\IncMargin{0.5em}


\begin{algorithm}[h]
    \small
    \DontPrintSemicolon
    \SetKwInOut{Input}{input}
    \SetKwInOut{Output}{output}
    \SetKwBlock{NoBegin}{}{end}





    \SetKwData{Valg}{$V$}
    \SetKwData{Ey}{$E_{y}$}
    \SetKwData{xalg}{$x$}
    
	\SetKwData{yalg}{$y$}
	\SetKwData{ralg}{$r$}
	\SetKwData{sigmaalg}{$\sigma$}
    \SetKwData{vx}{$v_{x}$}
    \SetKwData{ey}{$e_{y}$}
    \SetKwData{gvalue}{$\mathtt{g}$}
	\SetKwData{gbarvalue}{$\mathtt{\bar{g}}$}
	\SetKwData{hvalue}{$\mathtt{h}$}
	\SetKwData{vVYFrom}{$v_{y,\mathrm{from}}$}
	\SetKwData{vVYTo}{$v_{y,\mathrm{to}}$}
	\SetKwData{vVXFrom}{$v_{x,\mathrm{from}}$}
	\SetKwData{vVXTo}{$v_{x,\mathrm{to}}$}
	\SetKwData{vVSigmaFrom}{$v_{\sigma,\mathrm{from}}$}
	\SetKwData{vVSigmaTo}{$v_{\sigma,\mathrm{to}}$}
	\SetKwData{vEX}{$e_{x}$}
	\SetKwData{psigma}{$\mathtt{p}_{\sigma}$}
	\SetKwData{py}{$\mathtt{p}_{y}$}
	\SetKwData{vy}{$v_{y}$}
	\SetKwData{vsigma}{$v_{\sigma}$}
	\SetKwData{ey}{$e_{y}$}
	\SetKwData{esigma}{$e_{\sigma}$}
	\SetKwData{tailalg}{$\mathtt{tail}$}
	\SetKwData{headalg}{$\mathtt{head}$}
	\SetKwData{outgoingalg}{$\mathtt{outgoing}$}
	
    \setlength{\columnsep}{2mm}
	\begin{multicols*}{2}
	\vspace*{-4mm}
	\SetFuncSty{textbf}
    \fInitOutputNode{$\yalg$}
    \SetFuncSty{texttt}
    \NoBegin
    {
	    $\vy.\yalg \leftarrow \yalg$;
	    
	    $\vy.\gvalue \leftarrow \infty$;
	    $\vy.\gbarvalue  \leftarrow \infty$;
	    
	    $\vy.\hvalue  \leftarrow 0$;
	    

	    
		$\vy.\py \leftarrow \varnothing$;
	    $\vy.\psigma \leftarrow \varnothing$;
	            

    	
    	\Return{$\vy$};
    }

    \SetFuncSty{textbf}
    \fInitOutputEdge{$\vVYFrom,\vVYTo,\ralg$}
    \SetFuncSty{texttt}
    \NoBegin
    {
	    $\ey.\tailalg \leftarrow \vVYFrom$;
	    
	    $\ey.\headalg \leftarrow \vVYTo$;
	    	
	    $\ey.\ralg \leftarrow \ralg$;
	            
    	\Return{$\ey$};
    }
    
    \SetFuncSty{textbf}
    \fInitTrajectoryNode{$\sigmaalg,\ey, \Ey$}
    \SetFuncSty{texttt}
    \NoBegin
    {
	    $\vsigma.\sigmaalg \leftarrow \sigmaalg$;
	    
	    $\vsigma.\ey \leftarrow \ey$;
	    
	    $\vsigma.\outgoingalg \leftarrow \Ey$;
	    


    	
    	\Return{$\vsigma$};
    }
    
    \vspace*{-4mm}
    \SetFuncSty{textbf}
    \fInitTrajectoryEdge{$\vVSigmaFrom,\vVSigmaTo, \sigmaalg$}
    \SetFuncSty{texttt}
    \NoBegin
    {
	    $\esigma.\tailalg \leftarrow \vVSigmaFrom$;
	    
	    $\esigma.\headalg \leftarrow \vVSigmaTo$;
	    
	    $\esigma.\sigmaalg \leftarrow \sigmaalg$;
	    
       	\Return{$\esigma$};
    }
        
    \SetFuncSty{textbf}
    \fInitStateNode{$\xalg$}
    \SetFuncSty{texttt}
    \NoBegin
    {
	    $\vx.\xalg \leftarrow \xalg$;
       	
       	\Return{$\vx$};
    }
    
    \SetFuncSty{textbf}
    \fInitStateEdge{$\vVXFrom,\vVXTo,\sigmaalg$}
    \SetFuncSty{texttt}
    \NoBegin
    {
	    $\vEX.\tailalg \leftarrow \vVXFrom$;
    
	    $\vEX.\headalg \leftarrow \vVXTo$;
	    
	    $\vEX.\sigmaalg \leftarrow \sigmaalg$;
          	
        \Return{$\vEX$};
    }
    
	\end{multicols*}
	\vspace*{2mm}
\caption{{\small Node and Edge Constructor Procedures}{\color{white}$^\#$}} \label{alg:constructor_procedures_closed_loop_rrtsharp}

\end{algorithm}
\DecMargin{0.5em} 

\vspace*{-5pt}
}

\subsection{Properties of the Algorithm}

The \AlgClosedLoopRRTsharp{} algorithm provides both dynamic feasibility guarantees, that is, the lowest-cost reference trajectory computed by the algorithm can be tracked by the low-level controller, and asymptotic optimality guarantees, that is, the lowest-cost reference trajectory computed by the algorithm converges to the optimal reference trajectory almost surely. The former property is an immediate result of using closed-loop prediction during the search phase. During the extension of the graph $\vClGY$, if some segments of a reference trajectory can not be tracked, that is, is not dynamically feasible, the corresponding state trajectory is not stored in the graph $\vClGSigma$ constructed by the algorithm. The former property is due to the asymptotic optimality property of the \AlgRRTsharp{} algorithm~\cite{arslan2013useofrelaxation}. The proposed algorithm incrementally grows a graph $\vClGY$ in the output space in a similar fashion as the \AlgRRG{} algorithm does~\cite{karaman2010optimal}. Therefore, the lowest-cost path encoded in $\vClGY$ converges to the optimal output trajectory in the output space almost surely. In addition, the lowest-cost output trajectory encoded in the graph $\vClGY$ is extracted at the end of each iteration in a similar fashion as the \AlgRRTsharp{} algorithm does. Given the cost function that associates each edge in $\vClGY$ with a non-negative cost values being \textit{monotonic} and \textit{bounded}, the proposed algorithm is asymptotically optimal.


\section{Numerical Study}\label{section:numerical_simulations}
The proposed algorithm is evaluated on two scenarios where a nonholonomic, wheeled vehicle, modeled as a unicycle, travels along a track. The motion equations are
\begin{align*}
\dot{x}_{1} &= x_{4} \sin (x_{3}), \;
\dot{x}_{2} = x_{4} \cos (x_{3}),\;
\dot{x}_{3} = u_{1}, \;
\dot{x}_{4} = u_{2},   \\
y_{1} &= x_{1}, \; 
y_{2} = x_{2},
\end{align*}
where $x_{1}$, $x_{2}$ are the Cartesian coordinates  of the vehicle, $x_{3}$ is the  heading angle, $x_{4}$ is the translational velocity , and $u_{1}$, $u_{2}$ are the controls for the angular and translational velocity. Each control input takes values in an interval, that is, $u_{i} \in [u^{l}_{i}, u^{u}_{i}]$. A pure-pursuit controller  tracks a given reference path~\cite{Amidi1990229}. The heading command is generated by following a look-ahead point on a given reference path. The speed command is given as a desired  speed $v_{\mathrm{crs}}$, which  is tracked by a proportional controller.

\begin{figure*}
\centering
	\mbox{
    \subfigure[]{\scalebox{0.3}{\includegraphics[trim = 4.0cm 6.937cm 3.587cm 7.0cm, clip =
          true]{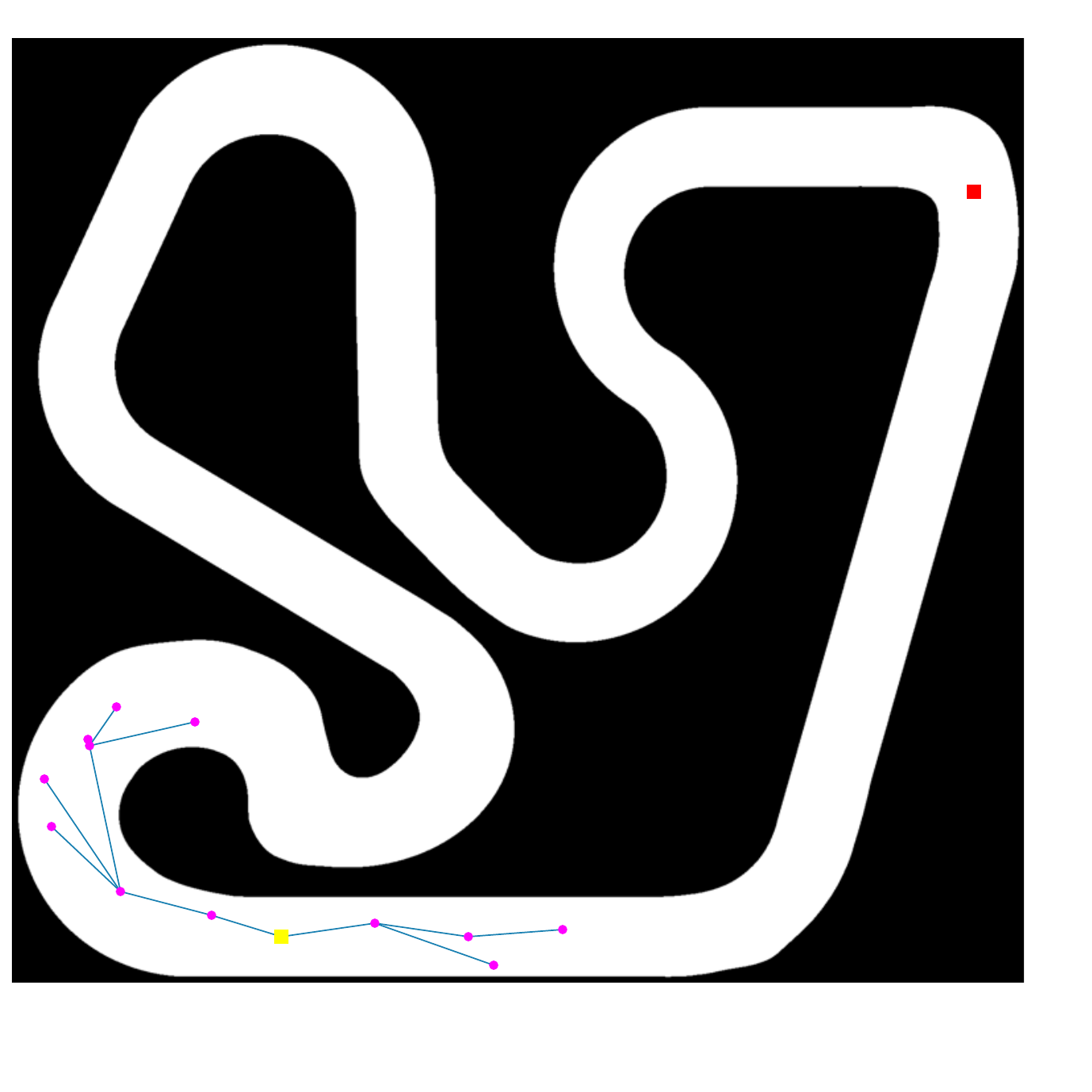}} \label{figure:pt1_reference_closed_loop_rrtsharp_it00050}}
    \subfigure[]{\scalebox{0.3}{\includegraphics[trim = 4.0cm 6.937cm 3.587cm 7.0cm, clip =
          true]{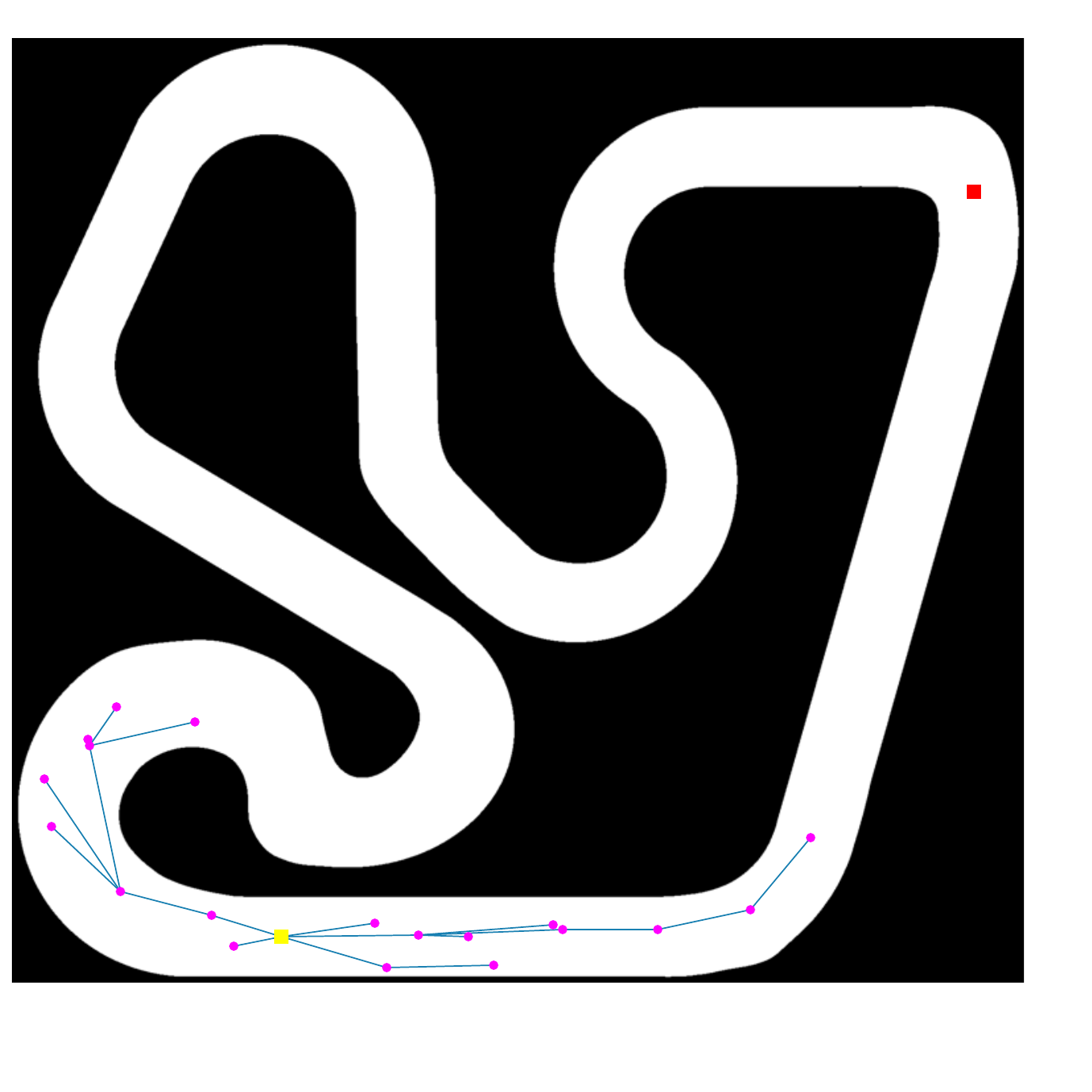}} \label{figure:pt1_reference_closed_loop_rrtsharp_it00100}}
    \subfigure[]{\scalebox{0.3}{\includegraphics[trim = 4.0cm 6.937cm 3.587cm 7.0cm, clip =
          true]{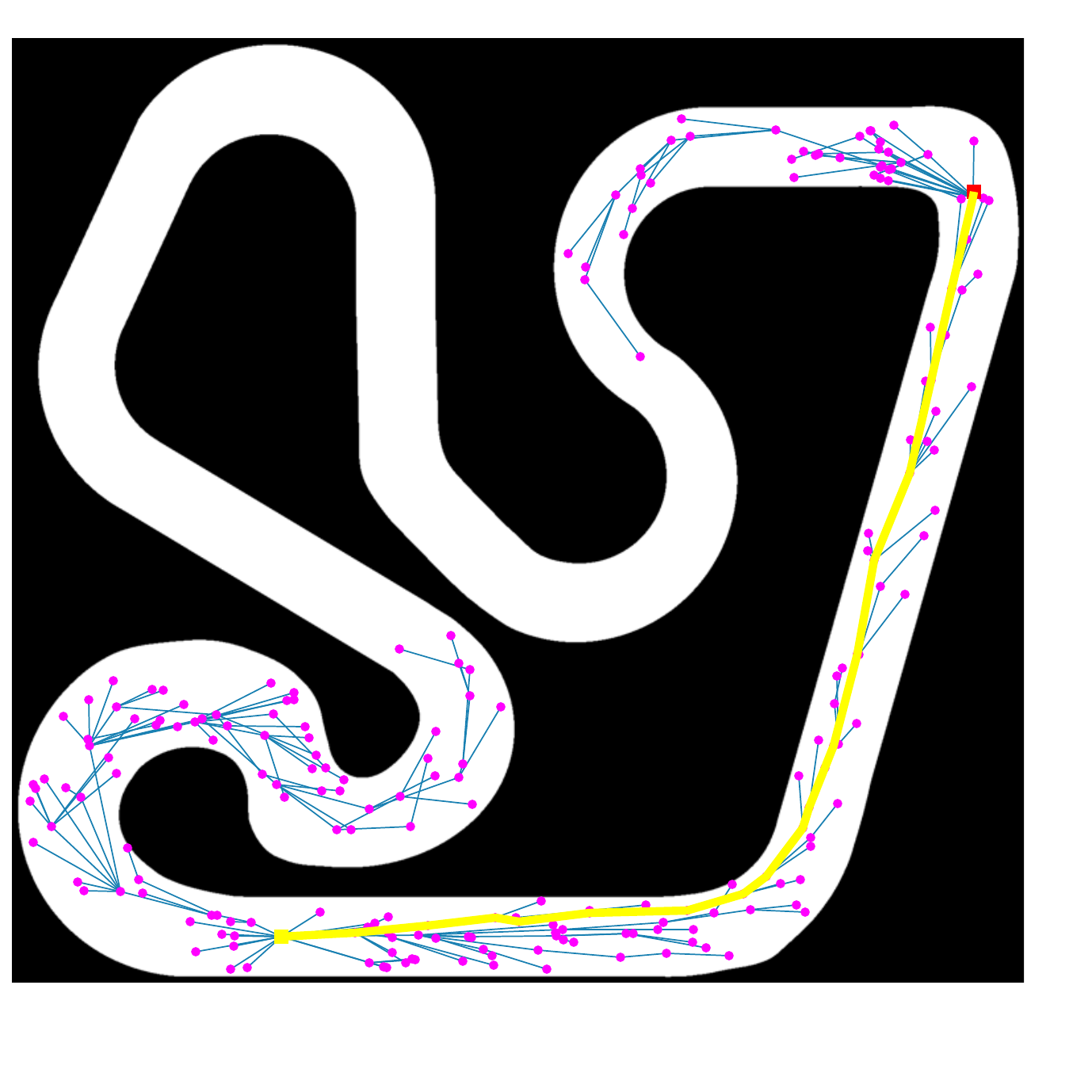}} \label{figure:pt1_reference_closed_loop_rrtsharp_it00500}}
    \subfigure[]{\scalebox{0.3}{\includegraphics[trim = 4.0cm 6.937cm 3.587cm 7.0cm, clip =
          true]{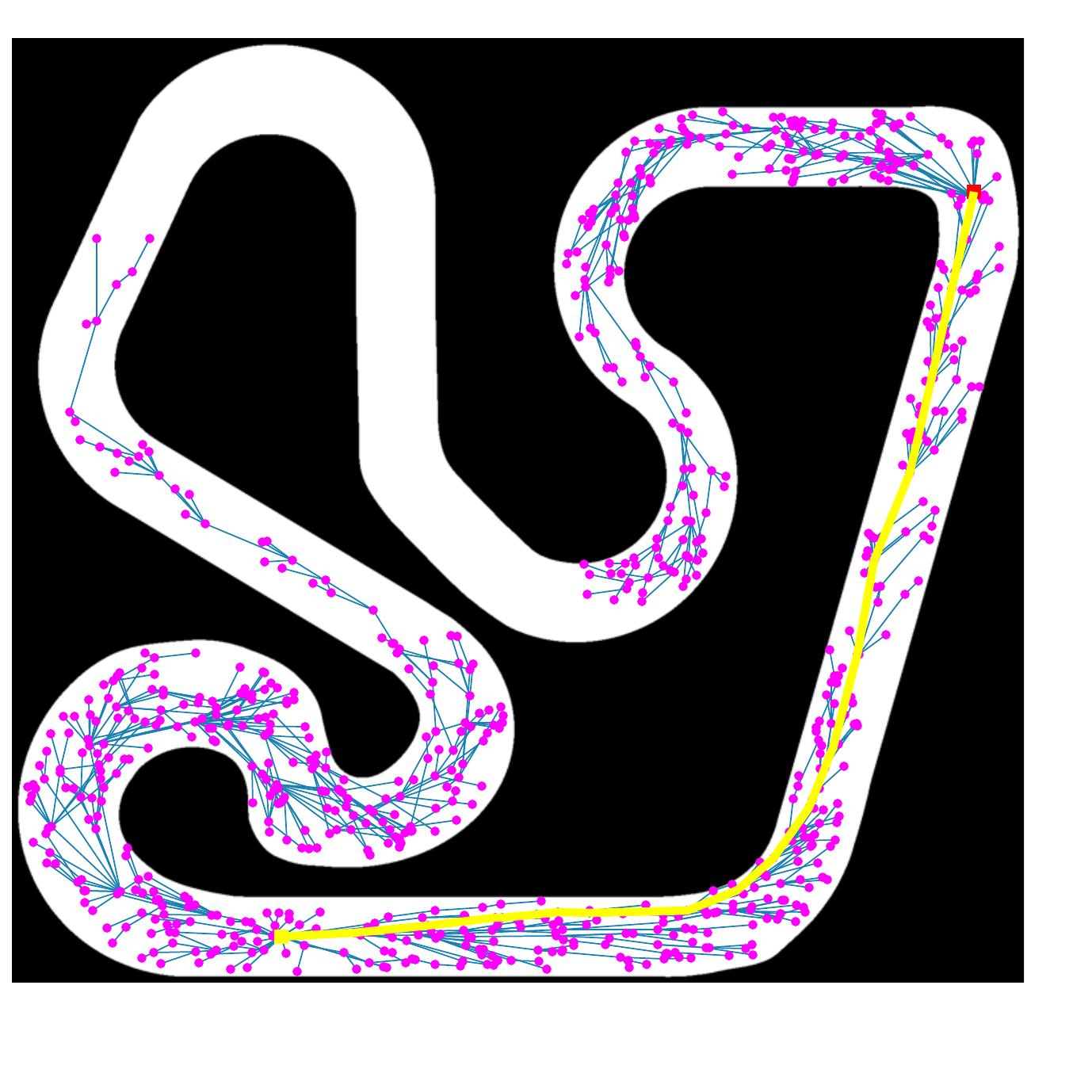}} \label{figure:pt1_reference_closed_loop_rrtsharp_it01500}}
 }

	\mbox{
    \subfigure[]{\scalebox{0.3}{\includegraphics[trim = 4.0cm 6.937cm 3.587cm 7.0cm, clip =
          true]{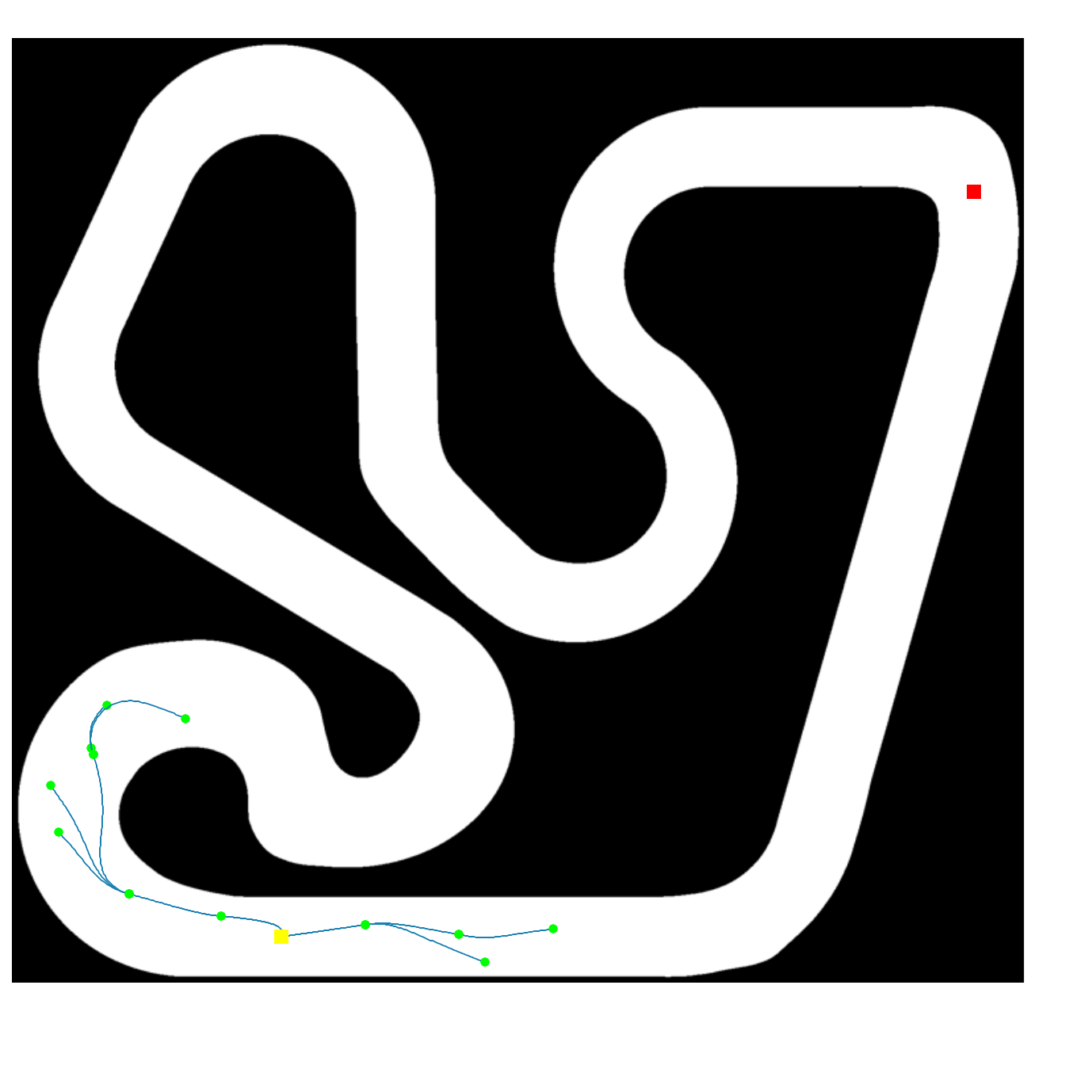}} \label{figure:pt1_state_closed_loop_rrtsharp_it00050}}
    \subfigure[]{\scalebox{0.3}{\includegraphics[trim = 4.0cm 6.937cm 3.587cm 7.0cm, clip =
          true]{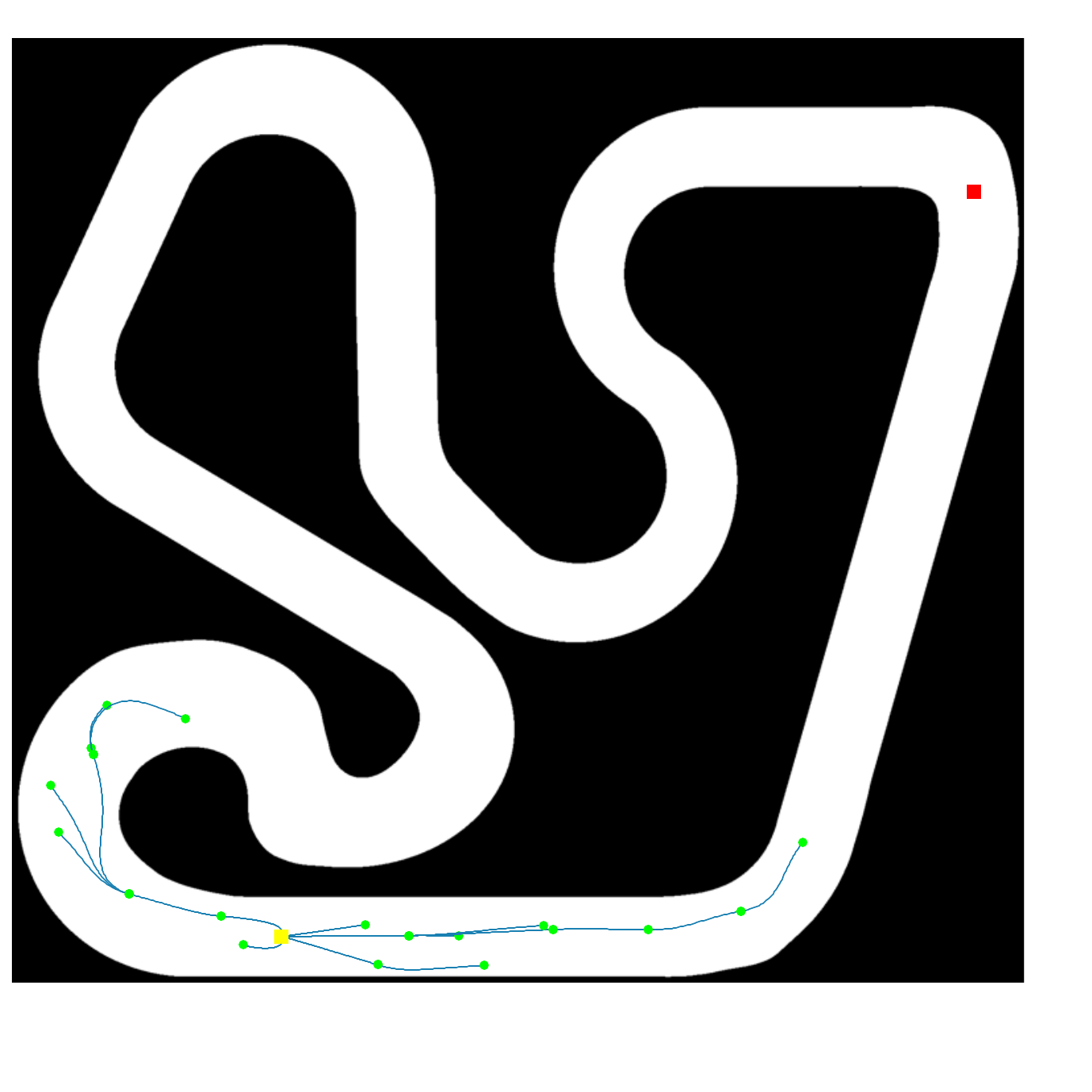}} \label{figure:pt1_state_closed_loop_rrtsharp_it00100}}
    \subfigure[]{\scalebox{0.3}{\includegraphics[trim = 4.0cm 6.937cm 3.587cm 7.0cm, clip =
          true]{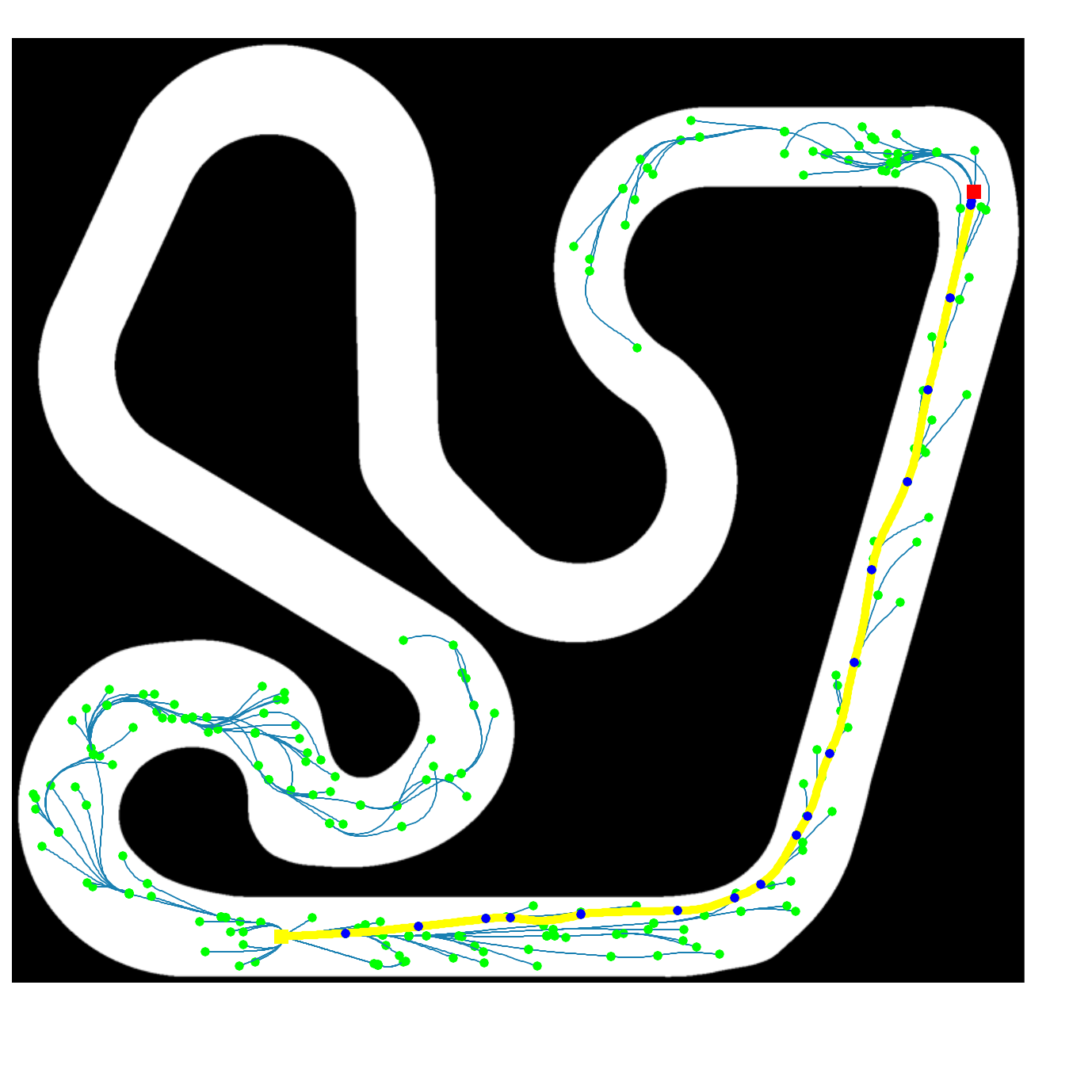}} \label{figure:pt1_state_closed_loop_rrtsharp_it00500}}
    \subfigure[]{\scalebox{0.3}{\includegraphics[trim = 4.0cm 6.937cm 3.587cm 7.0cm, clip =
          true]{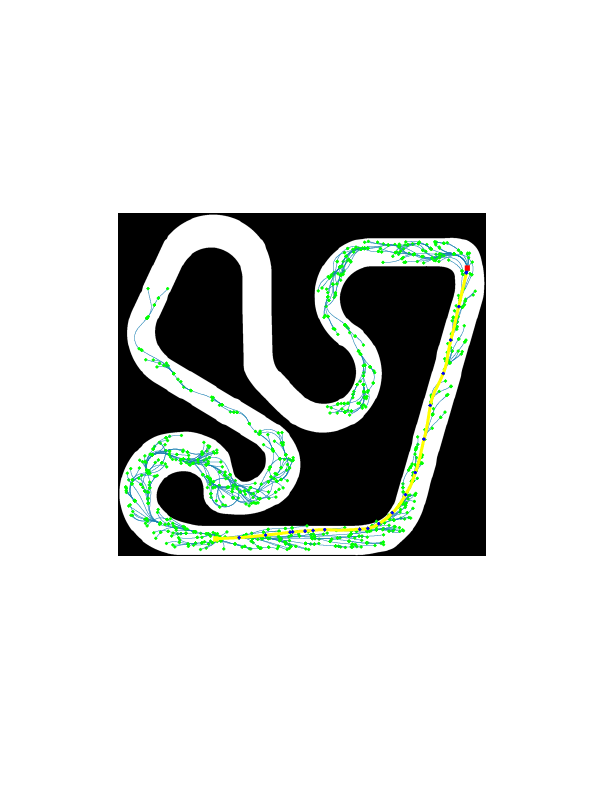}} \label{figure:pt1_state_closed_loop_rrtsharp_it01500}}
    }

    \caption{The evolution of the solution trees for reference paths and state trajectories computed by \AlgClosedLoopRRTsharp{} are shown in  \subref{figure:pt1_reference_closed_loop_rrtsharp_it00050}-\subref{figure:pt1_reference_closed_loop_rrtsharp_it01500} and \subref{figure:pt1_state_closed_loop_rrtsharp_it00050}-\subref{figure:pt1_state_closed_loop_rrtsharp_it01500}, respectively. The trees \subref{figure:pt1_reference_closed_loop_rrtsharp_it00050}, \subref{figure:pt1_state_closed_loop_rrtsharp_it00050} are at 50 iterations, \subref{figure:pt1_reference_closed_loop_rrtsharp_it00100}, \subref{figure:pt1_state_closed_loop_rrtsharp_it00100} are at 100 iterations, \subref{figure:pt1_reference_closed_loop_rrtsharp_it00500}, \subref{figure:pt1_state_closed_loop_rrtsharp_it00500} are at 500 iterations,
     and \subref{figure:pt1_reference_closed_loop_rrtsharp_it01500}, \subref{figure:pt1_state_closed_loop_rrtsharp_it01500} are at 1500 iterations.
}
     \label{figure:sim_pt1}
\end{figure*}

First, the objective is point-to-point navigation in the counter-clockwise direction on a race track, while minimizing the Euclidean path length. The track size is (100m$\times$100m) and the origin is located at its center.  \AlgClosedLoopRRTsharp  executed for 1,500 iterations. Fig.~\ref{figure:sim_pt1} shows the resulting tree at different stages. Initially, the vehicle is at $(-25,-45)$, with zero heading angle and  zero speed (yellow square at bottom-left). The task is to move to  $(48,33)$ (red square at top-right). As seen in Figs.~\ref{figure:sim_pt1}\subref{figure:pt1_reference_closed_loop_rrtsharp_it00050}-\subref{figure:pt1_reference_closed_loop_rrtsharp_it01500}, the algorithm incrementally grows a graph in the output space $(x_{1},x_{2})$. Each path in the graph corresponds to a  reference path, used as an input to the closed-loop system.  \AlgClosedLoopRRTsharp  quickly computes a long reference path. Then, it seeks alternative paths of the graph as more information is explored and improves the existing solution if closed-loop simulation of a new reference path yields lower cost. The nodes and edges of the graph correspond to waypoints and straight line segments. The lowest-cost path is shown in yellow. The value is 127.2. Figs.~\ref{figure:sim_pt1}\subref{figure:pt1_state_closed_loop_rrtsharp_it00050}-\subref{figure:pt1_state_closed_loop_rrtsharp_it01500} shows the  state trajectories, computed during closed-loop simulation in \AlgClosedLoopRRTsharp. 


\begin{figure*}
\centering
	\mbox{
    \subfigure[]{\scalebox{0.3}{\includegraphics[trim = 4.0cm 6.937cm 3.587cm 7.0cm, clip =
          true]{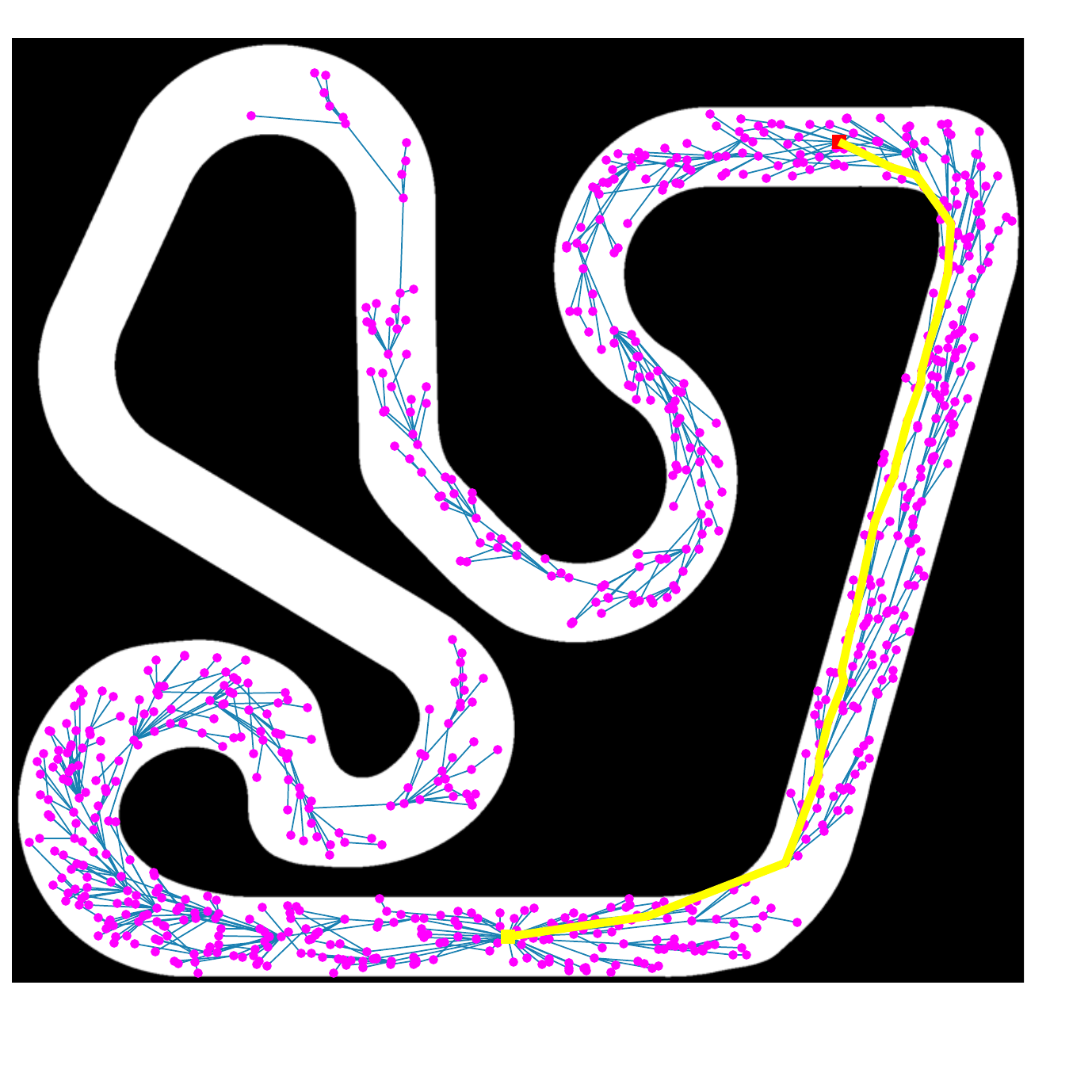}} \label{figure:pt1_reference_closed_loop_rrtsharp_it00050}}
    \subfigure[]{\scalebox{0.3}{\includegraphics[trim = 4.0cm 6.937cm 3.587cm 7.0cm, clip =
          true]{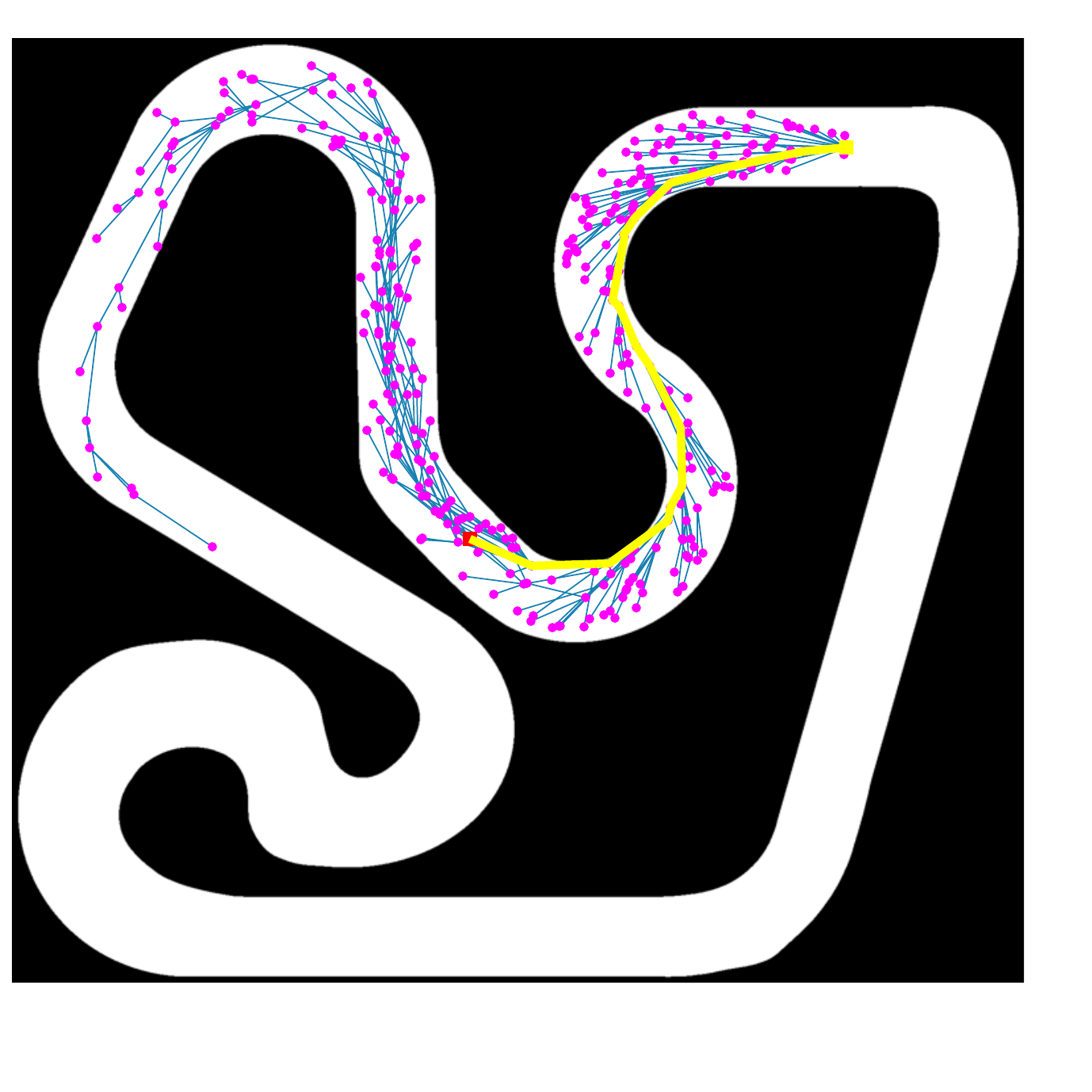}} \label{figure:pt1_reference_closed_loop_rrtsharp_it00100}}
    \subfigure[]{\scalebox{0.3}{\includegraphics[trim = 4.0cm 6.937cm 3.587cm 7.0cm, clip =
          true]{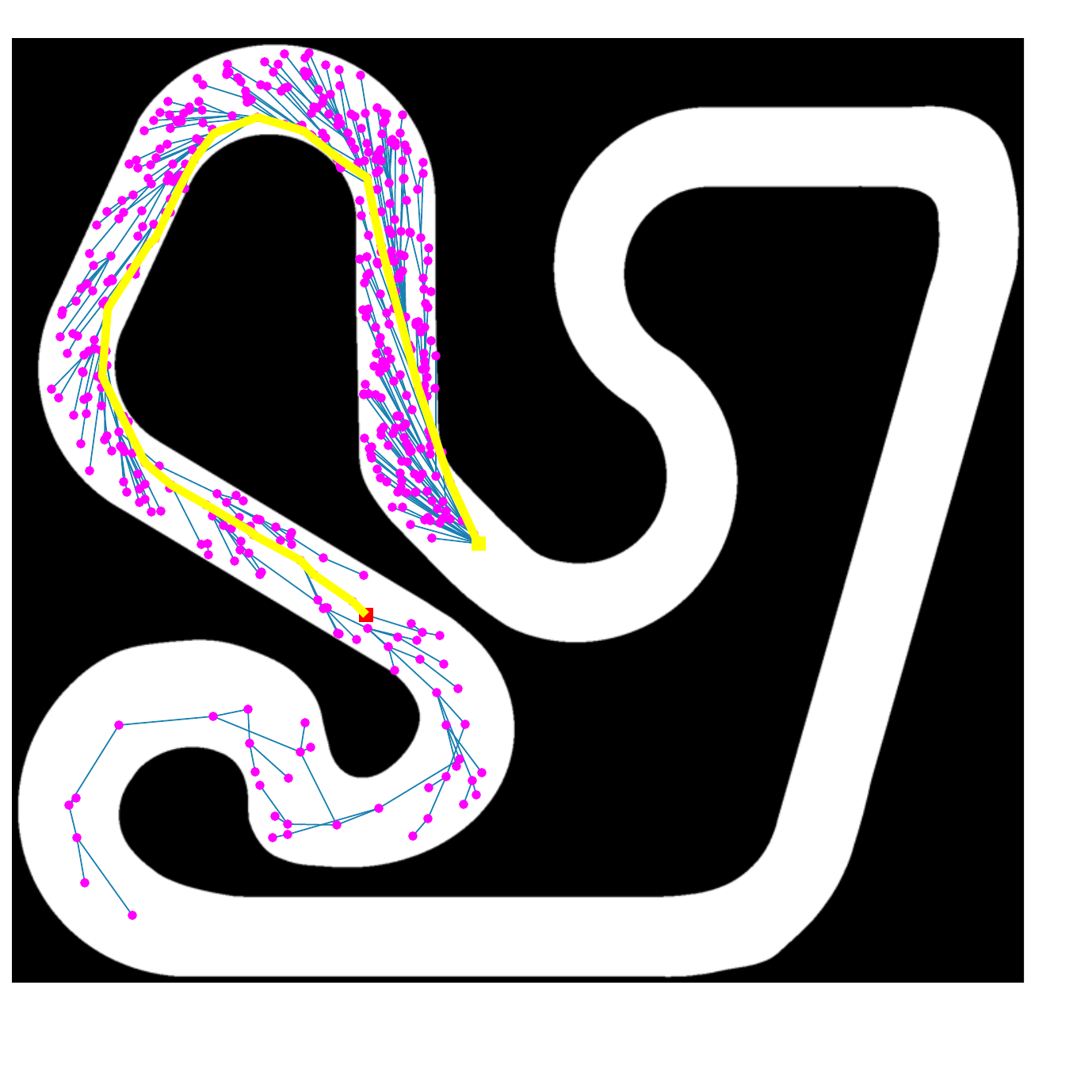}} \label{figure:pt1_reference_closed_loop_rrtsharp_it00500}}
    \subfigure[]{\scalebox{0.3}{\includegraphics[trim = 4.0cm 6.937cm 3.587cm 7.0cm, clip =
          true]{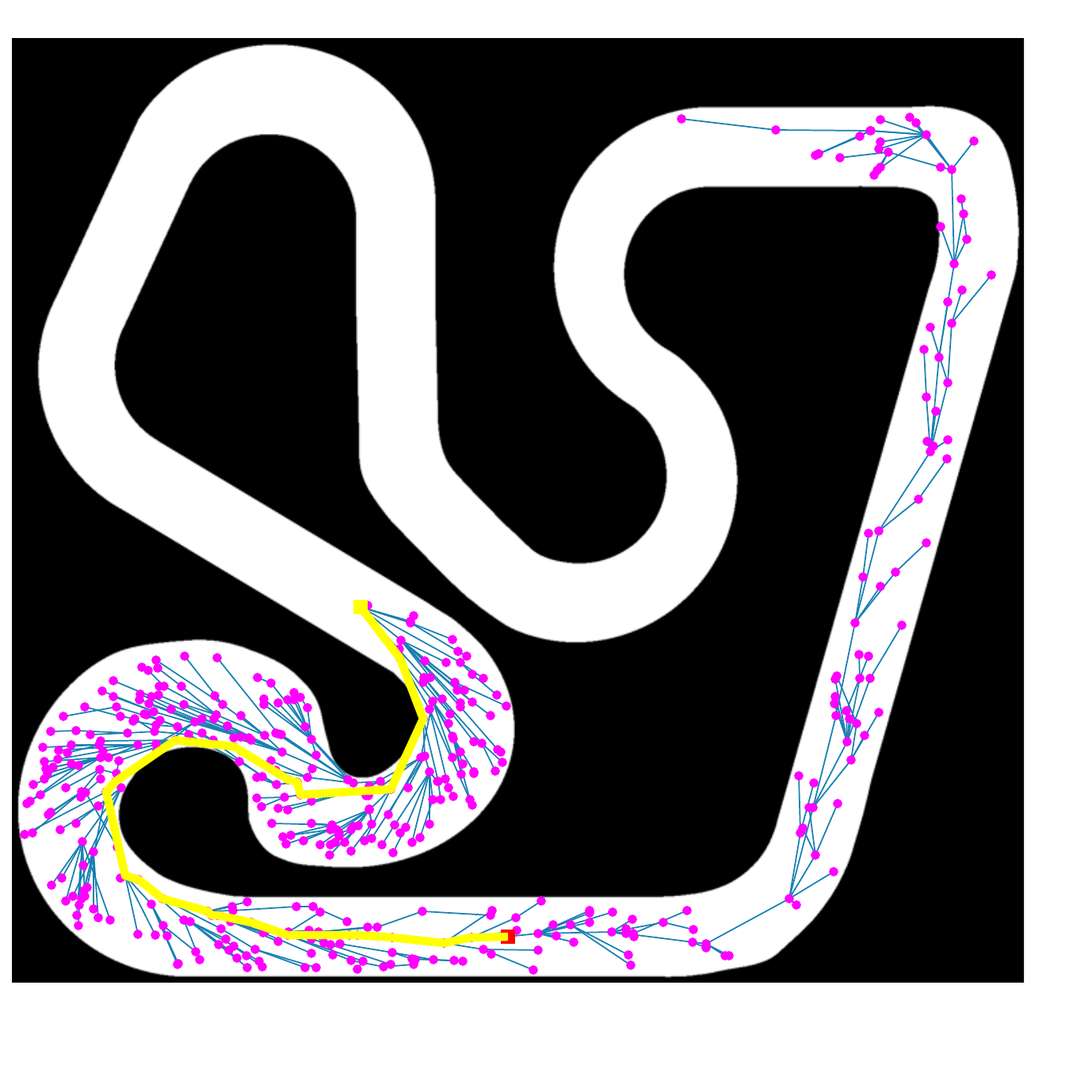}} \label{figure:pt1_reference_closed_loop_rrtsharp_it01500}}
 }

	\mbox{
    \subfigure[]{\scalebox{0.3}{\includegraphics[trim = 4.0cm 6.937cm 3.587cm 7.0cm, clip =
          true]{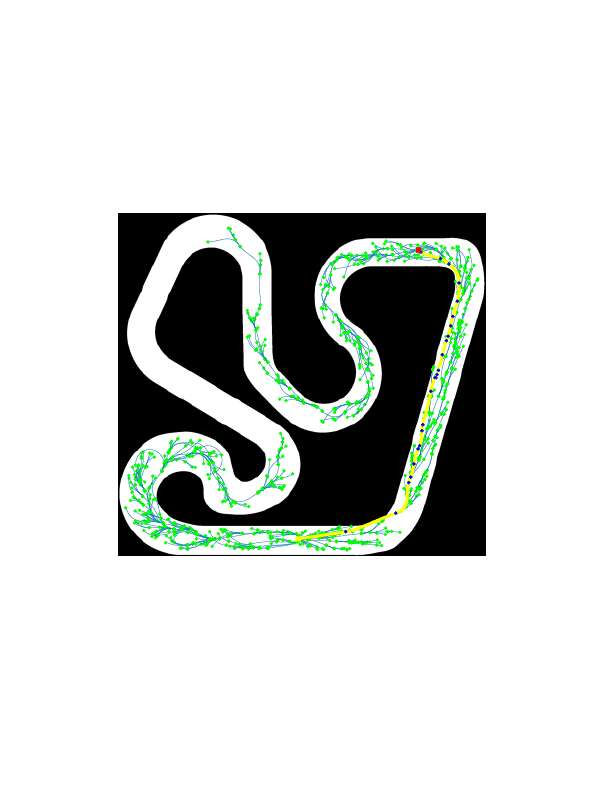}} \label{figure:pt1_state_closed_loop_rrtsharp_it00050}}
    \subfigure[]{\scalebox{0.3}{\includegraphics[trim = 4.0cm 6.937cm 3.587cm 7.0cm, clip =
          true]{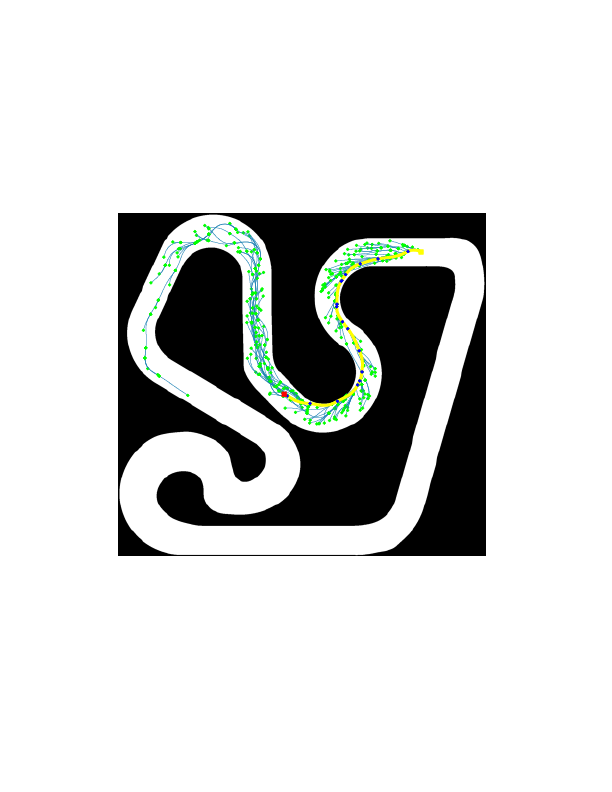}} \label{figure:pt1_state_closed_loop_rrtsharp_it00100}}
    \subfigure[]{\scalebox{0.3}{\includegraphics[trim = 4.0cm 6.937cm 3.587cm 7.0cm, clip =
          true]{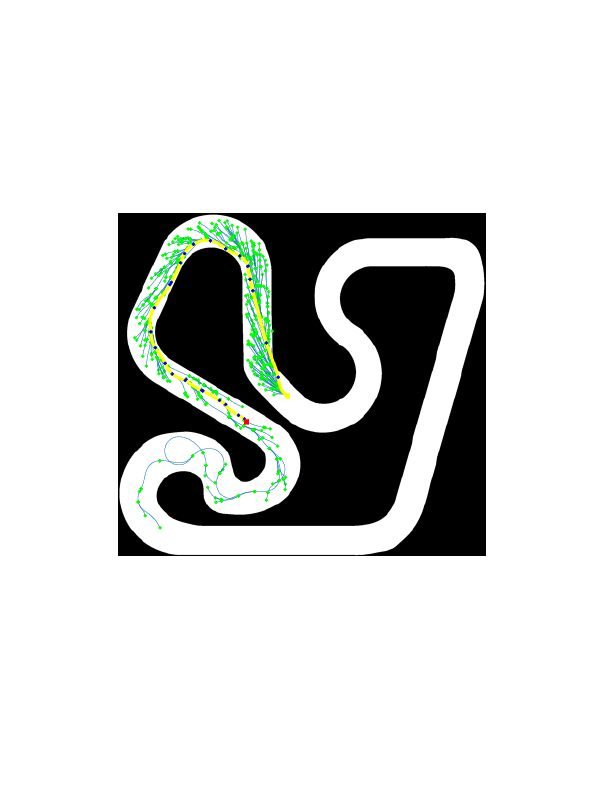}} \label{figure:pt1_state_closed_loop_rrtsharp_it00500}}
    \subfigure[]{\scalebox{0.3}{\includegraphics[trim = 4.0cm 6.937cm 3.587cm 7.0cm, clip =
          true]{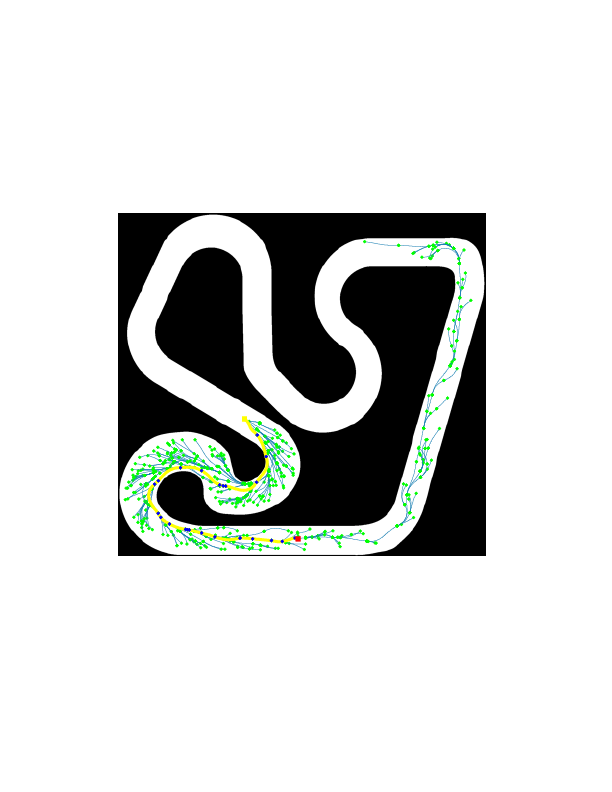}} \label{figure:pt1_state_closed_loop_rrtsharp_it01500}}
    }

    \caption{Results from a simulation where the vehicle navigates four consecutive waypoints, given by a high-level navigator. The evolution of the trees for reference paths and state trajectories computed by \AlgClosedLoopRRTsharp{} are shown in  \subref{figure:pt1_reference_closed_loop_rrtsharp_it00050}-\subref{figure:pt1_reference_closed_loop_rrtsharp_it01500} and \subref{figure:pt1_state_closed_loop_rrtsharp_it00050}-\subref{figure:pt1_state_closed_loop_rrtsharp_it01500}, respectively. In each stage, 1,500 iterations are made. As the vehicle gets close to the current waypoint, the next waypoint is sent to the motion planner, similar to \cite{kuwata2008motion}.}
     \label{figure:sim_pt2}
\end{figure*}

In the second scenario, the goal is to recursively navigate the vehicle on the race track. The vehicle is tasked to navigate sequentially to a set of waypoints, presumably coming from a high-level navigator. In each stage, the \AlgClosedLoopRRTsharp algorithm was executed for 1,500 iterations to find a motion plan from the current state of the vehicle to a desired next waypoint. Each next waypoint is sent to the motion planner as the vehicle gets close to the current waypoint, similar to \cite{kuwata2008motion}. In this simulation, the vehicle is tasked to navigate four waypoints sequentially. The solution trees of reference paths and corresponding state trajectories for each step are shown in Fig.~\ref{figure:sim_pt2}. As seen during simulations, leveraging the dynamics information of the vehicle during the search phase allows to construct dynamically feasible paths and avoid shortest paths that pass close to the boundary of the track.


\section{Conclusion}\label{section:conclusion}

We presented  a new asymptotically optimal motion-planning algorithm, called \AlgClosedLoopRRTsharp{}, using closed-loop prediction for trajectory generation. The approach is a hybrid of the \AlgClosedLoopRRT{} and the \AlgRRTsharp{} algorithms. It incrementally grows a graph of reference trajectories, used as inputs to a low-level tracking controller, and chooses the one that yields the lowest-cost state trajectory of the closed-loop system. \AlgClosedLoopRRTsharp{} provides dynamic feasibility by construction and ensures asymptotic optimality, that is, it finds the optimal reference trajectory given controller. Simulation results on a nonholonomic system showed the efficacy of the approach.

\bibliography{arxiv}
\bibliographystyle{plain}

\end{document}